\documentclass[10pt,twocolumn,letterpaper]{article}

\usepackage[pagenumbers]{cvpr} % To force page numbers, e.g. for an arXiv version
\usepackage[dvipsnames]{xcolor}

\definecolor{cvprblue}{rgb}{0.21,0.49,0.74}
\definecolor{greenwords}{RGB}{106,168,79} % RGB values for the color
\usepackage[pagebackref,breaklinks,colorlinks,citecolor=cvprblue]{hyperref}
\usepackage{array}
\usepackage{tabularx}
\usepackage{multirow}
\usepackage{color}
\usepackage[accsupp]{axessibility} % Improves PDF readability for those with visual impairments.
\def\refcls{RefExp${_\mathrm{cls}}$}
\def\refcandf{RefExp${_\mathrm{cand\_free}}$}

\def\paperID{4530} % *** Enter the Paper ID here
\def\confName{CVPR}
\def\confYear{2024}

\title{Enhancing Vision-Language Pre-training with Rich Supervisions}

\author{
Yuan Gao$^{1*{\dag}}$
\hspace{0.3cm}
Kunyu Shi$^{2*}$
\hspace{0.3cm}
Pengkai Zhu$^2$
\hspace{0.3cm}
Edouard Belval$^2$
\hspace{0.3cm}
Oren Nuriel$^2$
\hspace{0.3cm}
\\
Srikar Appalaraju$^2$
\hspace{0.3cm}
Shabnam Ghadar$^2$
\hspace{0.3cm}
Vijay Mahadevan$^2$
\hspace{0.3cm}
Zhuowen Tu$^2$
\hspace{0.3cm}
Stefano Soatto$^2$ \\[.5ex]
$^1$Stanford University \hspace{0.5cm}
$^2$AWS AI Labs \hspace{0.5cm} \\ [.5ex]
{\tt\small y1gao@stanford.edu} \\
{\tt\small \{kunyus,  zhpengka, belvae, onuriel, srikara, shabnam, ztu, soattos\}@amazon.com}
}

\begin{document}
\maketitle

\begin{abstract}
We propose Strongly Supervised pre-training with ScreenShots (S4) - a novel pre-training paradigm for Vision-Language Models using data from large-scale web screenshot rendering.
Using web screenshots unlocks a treasure trove of visual and textual cues that are not present in using image-text pairs. 
In S4, we leverage the inherent tree-structured hierarchy of HTML elements and the spatial localization to carefully design 10 pre-training tasks with large scale annotated data. These tasks resemble downstream tasks across different domains and the annotations are cheap to obtain. We demonstrate that, compared to current screenshot pre-training objectives, our innovative pre-training method significantly enhances performance of image-to-text model in nine varied and popular downstream tasks - up to 76.1\% improvements on Table Detection, and at least 1\% on Widget Captioning.

\end{abstract}

\def\thefootnote{*}\footnotetext{Equal contribution}
\def\thefootnote{\textdagger}\footnotetext{Work conducted during an internship at Amazon.}

\section{Introduction}
\label{sec:intro}
In recent years, there has been significant progress in Language Models (LMs) \cite{brown2020language,raffel2020exploring,chung2022scaling,ouyang2022training} and Vision Language Models (VLMs) \cite{wang2022ofa,radford2021learning,chen2022pali, Shi_2024_CVPR, Lu2019ViLBERTPT, li2020oscar, zhang2021vinvl, Jia2021ScalingUV, Radford2021LearningTV, Zhou2022ConditionalPL, Li2022BLIPBL, Appalaraju2023DocFormerv2LFAAAI, Zhou2021LearningTP, Biten_2022_CVPR, Wang2022ImageAA, Gao2021CLIPAdapterBV, Zhou2019UnifiedVP, zhang2023musketeer, Wang2021VLMoUV, Huang2021SeeingOO, Lu2022UnifiedIOAU, Zeng2021MultiGrainedVL, Hu2021ScalingUV, Yang2022VisionLanguagePW, Zhang2021VinVLMV, Zhang2022GLIPv2UL, Yao2021CPTCP, Xu2021E2EVLPEV,Zhuge2021KaleidoBERTVP,Du2022ASO, Yang2021CausalAF, Shu2022TestTimePT, Chen2022VLPAS, Dou2022CoarsetoFineVP, Appalaraju_2021_ICCV, Kamath2021MDETRM, Jin2021AGP, Wang2022OFAUA, Li2021GroundedLP, Zhang2022DINODW, Zeng2022SocraticMC} exhibiting strong zero-shot generalization and adaptability to a wide range of tasks. Though they may differ in architecture, data and task formulation, such foundational models predominantly rely on large-scale pre-training on giant corpora of web scraped data which serves as the source of generalization capability - C4 \cite{2019t5_C4dataset}, The Pile \cite{pile}, Laion 5B \cite{schuhmann2022laion}.

The pre-training of LMs and VLMs were mostly studied separately. For LMs, the inputs and outputs reside within a homogeneous space, and pre-training tasks that reconstruct inputs such as Masked Language Modeling (MLM) \cite{devlin2018bert, raffel2020exploring} and Casual Language Modeling (CLM) \cite{radford2018improving,brown2020language} have exhibited the capability of learning knowledge from large corpora of text extracted from web crawls, which translates well to downstream task performance. On the other hand, although the input reconstruction type of pre-training for VLMs have shown performance improvements in certain settings, in general they are less effective compared to what is observed in language domain \cite{kwon2022masked, chen2022pali} due to heterogeneity of vision tasks \cite{deng2009imagenet, lin2014microsoft, lazarow2020learning}. In addition to self-supervised learning tasks, many VLMs \cite{wang2022ofa,chen2022pali} use a mixture of supervised pre-training tasks (e.g. object detection, VQA, captioning, etc), relying on human manually labeled datasets such as COCO \cite{lin2014microsoft}, Object365 \cite{shao2019objects365}, VQA \cite{goyal2017making}, etc as well as datasets generated in automated fashion such as LAION-5B \cite{schuhmann2022laion}, WebLi-10B \cite{chen2022pali}.

\begin{figure}[!t]

\centering
\includegraphics[width=0.5\textwidth]{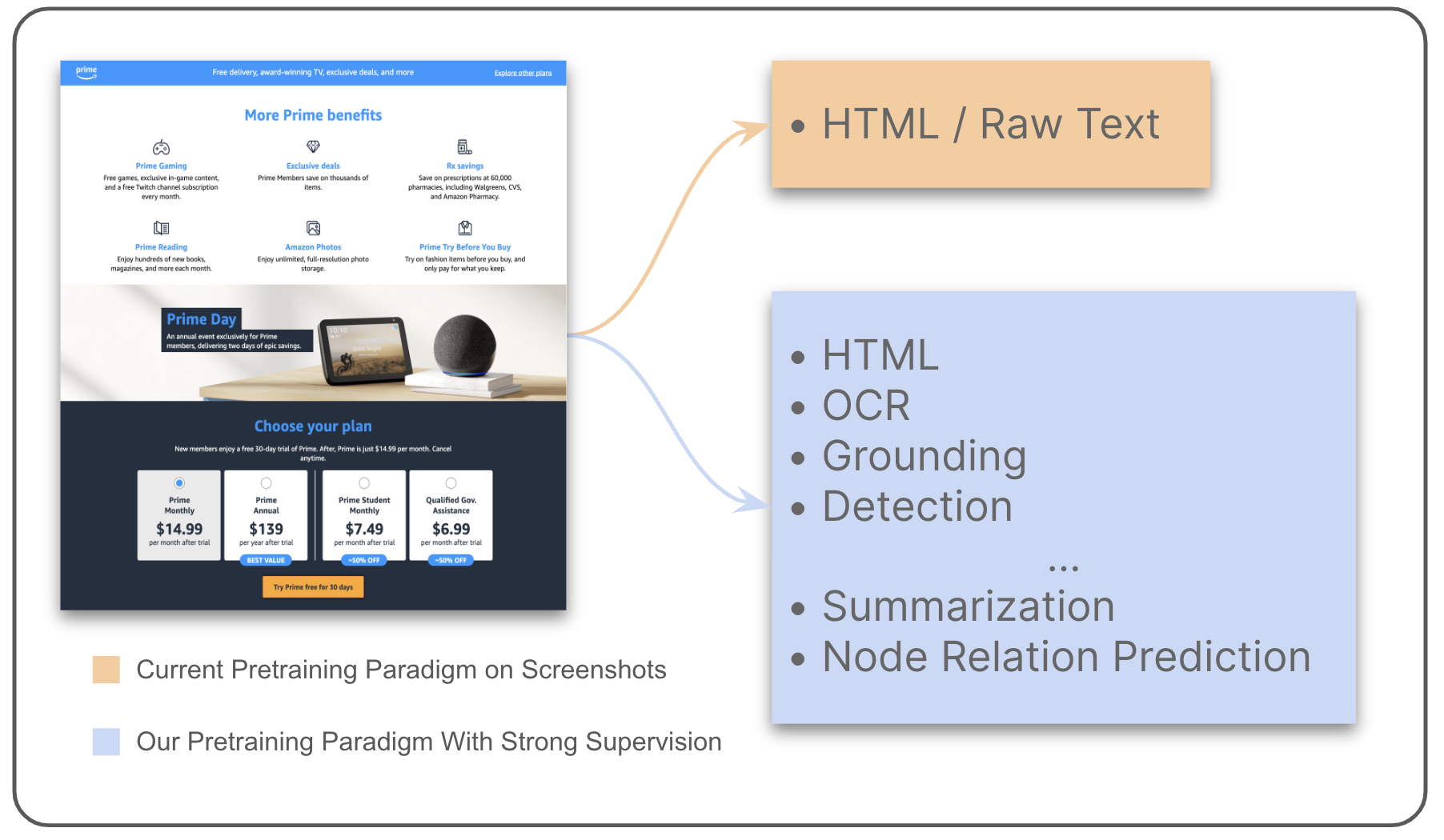}

\caption{We propose a novel pre-training paradigm - S4, composed of ten carefully designed tasks on large scale web-screenshots. Compared to image-to-text pretraining objectives on screenshots, which mainly utilized HTML\cite{lee2023pix2struct} or its subset like raw texts\cite{kim2022ocrfree,li2023spotlight}, our paradigm utilizes rich and diverse supervisions generated from web rendering that is also cheap to obtain.}
\label{fig:hist}
\vspace{-5mm}
\end{figure}

Advancements of supervised datasets have powered the advancements of VLMs. Increasing amounts of human annotated datasets were released \cite{kirillov2023segment, kuznetsova2020open}, which benefit the performance of similar or relevant downstream tasks, albeit at a steep cost. Approaches that can automatically generate supervisions at scale have also been explored \cite{schuhmann2022laion, lee2023pix2struct}. Notably, the use of massive amount of image-caption pairs, which are automatically generated using images and their associated Hypertext Markup Language (HTML) alt-text has enabled the development of some important VLMs such as CLIP models \cite{radford2021learning} and diffusion models \cite{rombach2022high}. Similarly, the use of screenshots and simplified HTML text pairs powered the Pix2Struct models \cite{lee2023pix2struct}. However, methods capable of producing automatically annotated data beyond basic image-text pairs are currently under explored. Consequently, the effects of employing explicit, automatically generated, fine-grained supervision for pre-training have been understudied. 

Therefore, in this work, we extend the use of web crawl corpuses and propose a novel pre-training framework that utilizes rich and diverse supervisions generated from web rendering. Modern websites are built using a combination of technologies such as HTML, CSS, JavaScript, that enable website creators to design dynamic content and interactive elements with diverse layouts. 

To leverage such information, our solution renders crawled web-pages into screenshot images. We also have access to textual content, position, attribute and relationship of HTML elements - all of which can be obtained cheaply and utilized in pre-training.
Building on this extracted data, we propose a set of pre-training tasks (see details in \ref{sec:pretrain_tasks}) that are highly synergistic to downstream tasks. Our results demonstrate significant performance improvements compared to the image-to-text pre-training baseline. On average, we observed an improvement of \textbf{+2.7\%} points across 5 datasets (ChartQA, RefExp, Widget Captioning, Screen Summarization and WebSRC) with language outputs, and a notable average increase of \textbf{+25.3\%} points on 4 datasets (PubLayNet, PubTables1M, RefExp candidate free and ICDAR 2019 modern) with localization outputs. See more in Tables \ref{tab:exp_table} and \ref{tab:exp_grounding}. Our key contributions: 
\begin{itemize}
    \item We develop an automatic data annotation pipeline that is able to render web crawls and create rich labels. Coupled with our carefully designed data cleaning process, we create a high-quality and large-scale vision language pre-training dataset.
    \item We propose a novel pre-training paradigm - S4, composed of ten carefully designed tasks on large scale web-screenshots 
    showing the effectiveness on a wide range of benchmarks.
    \item Comparing to current screenshot pre-training objectives, our innovative pre-training method significantly enhances performance of image-to-text model in nine varied and popular downstream tasks - up to 76.1\% improvements on Table Detection, and at least 1\% on Widget Captioning.
\end{itemize}

\section{Related Work}
\label{sec:Related Work}

Next, we discuss in detail the difference to previous pre-training approaches.

{\bf \noindent Masked Signal Modeling.} Pre-training through self-supervision has revolutionized the field of natural language processing (NLP). Pioneering models like BERT \cite{raffel2020exploring} and GPTs \cite{radford2018improving, brown2020language} demonstrated the profound impact of self-supervised learning in enhancing generalization across a variety of language tasks. The success in NLP spurred analogous research in computer vision, leading to innovations in Masked Image Modeling (MIM) with approaches such as BEiT \cite{bao2021beit}, SimMIM \cite{xie2022simmim}, and MAE \cite{he2022masked} that recover masked pixels or patches, and improvements on classic vision tasks such as classification, semantic segmentation, etc are observed. In the domain of Vision Language(VL), MLIM \cite{arici2021mlim} and MaskVLM \cite{kwon2022masked} propose to integrate MLM and MIM and conduct VL pretraining in a joint manner.

{\bf \noindent Supervised Pre-training} In supervised pre-training, image-caption pair annotations are generated on a large scale automatically from web crawls. This enables the training of models that generalize well in tasks like classification, retrieval, and captioning, as seen in works like CLIP, OFA, PaLi \cite{radford2021learning, wang2022ofa, chen2022pali}. Donut \cite{kim2022ocrfree} proposes a OCR-free model that relies on text reading pre-training of documents. SPOTLIGHT uses \cite{li2023spotlight} region to text pre-training task on website and UI datasets. Pix2Struct \cite{lee2023pix2struct} leverages screenshot and image pairs with a screen parsing pre-training task that converts webpage screenshots to HTML text. Our work proposes a pre-training paradigm that goes beyond image-text pairing type of tasks. We develop a suite of diverse, heterogeneous tasks specifically crafted to mirror the nature of downstream applications.

\section{S4 Pre-training}
In this section, we propose a novel pre-training paradigm for Vision-Language Models — Strongly Supervised pre-training with ScreenShots (S4) from large scale website rendering. We will first describe the creation procedure of our dataset for S4 pretraining, which we will call S4 Data, and then go through our proposed pre-training tasks enabled by our novel preprocessing method.

\begin{figure*}
\centering

\label{fig:wide_graphic}
\vspace{4mm}

\centering
\includegraphics[scale=0.53]{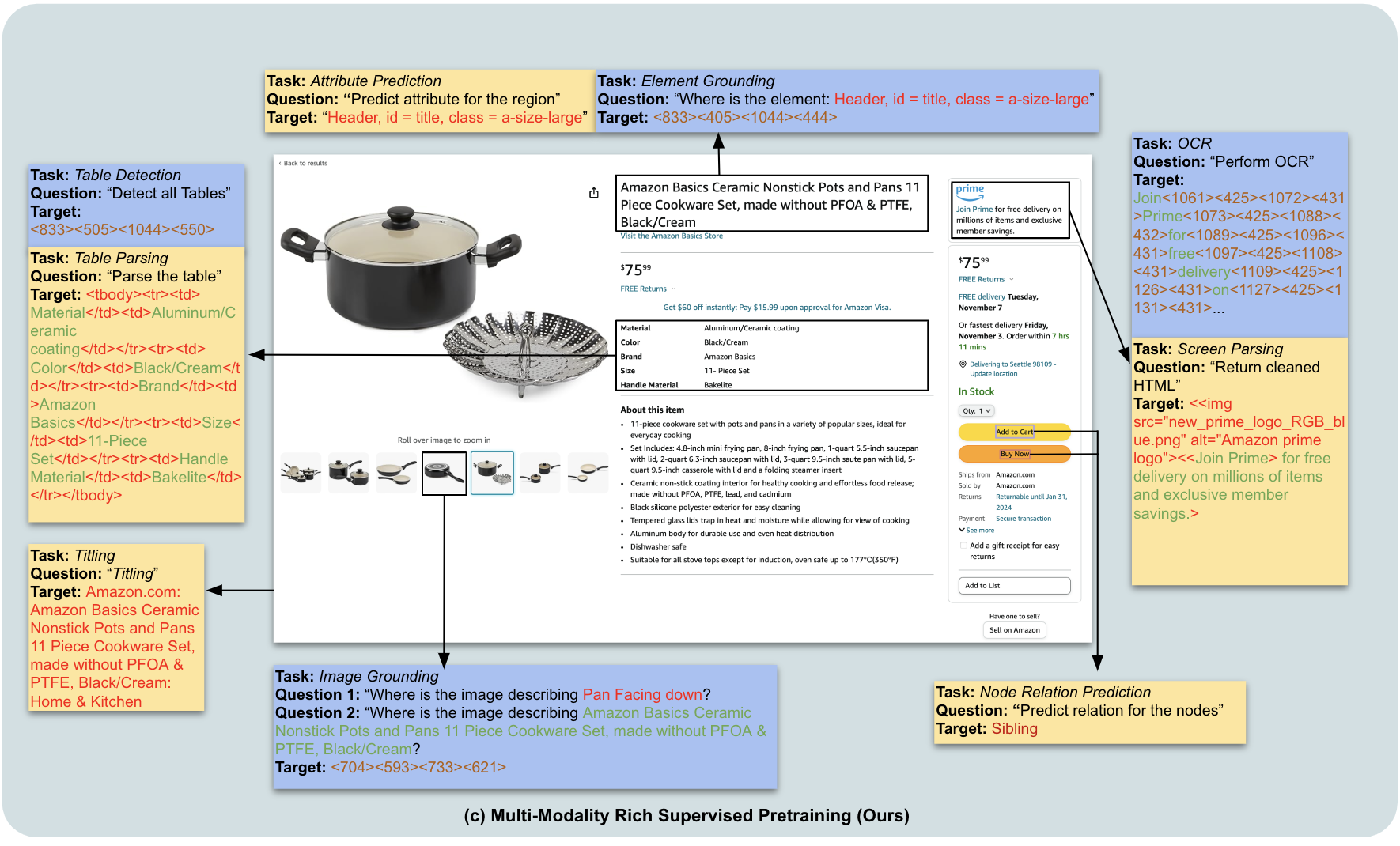}
\vspace{-0.5em}
\caption{Compared to traditional pre-training paradigms, our rich supervised pre-training leverages much more information that is also cheap to acquire (i.e via browser). We can then utilize the rich semantic and structural annotations to construct novel pre-training tasks that are naturally and directly aligned with downstream tasks. We use \textcolor{greenwords}{green} words to refer to the words contained (visible) in the screenshot. We use \textcolor{red}{red} words to refer to the words that are not visible in the screenshot. For instance, “price” is not shown on the screenshot, but is the id of an element (refer to picture). We use \textcolor{brown}{brown} words in the format of \texttt{<x><y><x><y>} to denote the bounding box.} 

\label{fig:wide_graphic}
\vspace{-1em}
\end{figure*}

\subsection{Dataset}
\subsubsection{Dataset Description}

CommonCrawl\footnote{http://commoncrawl.org/} provides access to a large-scale web page corpus spanning over a decade. We download the web crawls from the Registry of Open Data on AWS\footnote{https://registry.opendata.aws/commoncrawl/} and we filter content with an explicit copyright notice. We execute our rendering and extraction pipeline (described in \ref{sec:rendering_supervision}) and data pre-processing and cleaning procedure (described in \ref{sec:Cleaning_supervision}) to obtain 15M screenshots enriched with supervisions. We applied deduplication based on urls to make sure our screenshots are unique. Each page is rendered at a resolution of 1280x1280 and is paired with matching annotations that enable the proposed pre-training tasks described in \ref{sec:pretrain_tasks}.

\vspace{-0.5em}

\subsubsection{Efficient Rendering and Supervision Extraction}
\label{sec:rendering_supervision}

\vspace{-1em}

We use Playwright \footnote{https://github.com/microsoft/playwright}, which provides a programmatic interface to headless browsers that we use to render raw HTML files into screenshots. For each web page, we retrieve and cache the associated CSS, JavaScript fonts and images needed to render the page accurately. Caching those assets avoids making unnecessary requests to the originating website and quicker rendering, allowing us to create 5M parsed screenshot per day with 500 CPUs. 

We build the annotations by traversing through the document object model (DOM) tree and collecting annotations for every leaf node of type Text, Image, Table or Input. More information about the dataset and example annotation can be found in the supplementary material.

\subsubsection{Pre-processing and Cleaning}
\label{sec:Cleaning_supervision}
During data rendering, we found that directly traversing through the DOM tree and collecting information on each node would lead to the inclusion of elements that were not visible in the page. We solve this issue by inspecting their CSS property for visibility and verifying the alignment to their corresponding bounding box. Specifically, if the element \texttt{elem\_b} returned by clicking on the center \texttt{elem\_a}’s bounding box is not a descendent of \texttt{elem\_a}, then \texttt{elem\_a} is pruned. This simple heuristics helps us get rid of most of the annotation that contains invisible elements. Also, we implemented recursive pre-order traversal to filter out overflow words in a textnode where the texts are overflowing outside of it's ancestor's bounding box. Without such a filter, words that are occluded by other elements would be included in the final annotation. Finally, we get rid of all \texttt{<iframe>} tags since the Same Origin Policy prohibits direct access of the nodes in \texttt{<iframe>}.

\subsection{Pre-training Task construction}
\label{sec:pretrain_tasks}

Using the rich information provided by the HTML document structure, we design ten diverse supervised objectives, which improve upon previous supervision like Screen Parsing. Our tasks include: Screen Parsing,  OCR, Image Grounding, Element Grounding, Attribute Prediction, Node Relation Prediction, Table Detection, Table Parsing, Screen Titling, and Layout Analysis. We describe the objctives in the sections below, as well as in Figure \ref{fig:wide_graphic}.

\vspace{-1em}

\subsubsection{Screen Parsing}
Similar to Pix2Struct, our Screen Parsing objective aims to reconstruct both the texts and their underlying structure. As shown in Figure \ref{fig:wide_graphic}, the input is simply a screenshot with a bounding box drawn on a region, with words 50\% masked words, and the target is the cleaned HTML. We obtain the cleaned and simplified HTML as described in Pix2Struct by removing invisible nodes, and recursively remove chained nesting.

\vspace{-0.5em}

\subsubsection{Optical Character Recognition - OCR}
The OCR objective aims to train the model with the ability for spatial understanding. It takes in a screenshot with drawn bounding box specifying a region, and outputs a “\textcolor{greenwords}{word$_0$}\textcolor{brown}{\texttt{<x$_0$><y$_0$><x$_0$><y$_0$>}}\textcolor{greenwords}{word$_1$}\textcolor{brown}{\texttt{<x$_1$><y$_1$><x$_1$><y$_1$>}}...” sequence. To limit the sequence length, we choose bounding box for a random region that contains at most 50 words. The OCR objective empowers the model with the ability to spatially understand the screenshot, bringing benefits for general detection and grounding tasks.
\subsubsection{Image Grounding}
Image grounding is an important aspect to enhance image-text alignment, and we want to empower our model with the ability to understand the semantics of the images in the context of a screenshot. We formulate the Image grounding objective as follow: First, we randomly pick an \texttt{<img>} element from the screenshot. Then, we obtain two captions for the image: one is from the alt attribute, which we call \textcolor{red}{\texttt{alt\_caption}}, and second is from the text node that is closest to the \texttt{<img>} element in the HTML Tree, which we call \textcolor{greenwords}{\texttt{neighbor\_caption}}. We then randomly choose from \{\textcolor{red}{\texttt{alt\_caption}}, \textcolor{greenwords}{\texttt{neighbor\_caption}}\} and ask the model to predict the bounding box of the image described. Note that since the \textcolor{greenwords}{\texttt{neighbor\_caption}} appears in the screenshot, the model can, instead of learning the image-text relation, simply cheat to locate the \textcolor{greenwords}{\texttt{neighbor\_caption}} first and then predict the bounding box for the image closest to it. Therefore, to avoid leaking spatial information from texts, we mask out 90\% of the texts for the input screenshot to the model.

\vspace{-0.5em}

\subsubsection{Element Grounding}
\label{sec:element_grounding}
Element grounding is a generalization of image grounding to other elements in the HTML DOM tree. To build a better representation of the meaning and functionality of each elements shown in the screenshot, we ask the model to localize their position based on a text description. We obtain the text description by concatenating the element tag and attributes from \{class, id, label, for, alt, title, type\}. However, values of the attributes are often noisy as the id and class label of an element can be randomized (i.e, in web frontend frameworks such as React.js). We address this issue by adding a post-processing step that filters out words that are numerical, single characters or that combines letters and numbers, as they are unlikely to useful labels. As a final step we use the T5 tokenizer to get rid of strings that map to $<unk>$ tokens. 
\subsubsection{Attribute Prediction}
Beyond elements grounding from descriptions, we also ask the model to predict a matching description for a region in HTML. We group the visible elements into groups where they contain the same tag and attributes within \{class, id, label, for, alt, title, type\}, and randomly specify a group by rendering its bounding box to the input screenshot. The model is then asked to predict the tag and attributes in the following format: “\{tag\} \{tag.class\} \{tag.id\} \{tag.label\} \{tag.for\} \{tag.alt\}”. We apply the same post-processing described in \ref{sec:element_grounding} to filter out noise in the attribute values. The Attribute Prediction task forces the model to reason about the semantic meaning of each element, which could bring benefits downstreams tasks that involves element-level understanding.
\subsubsection{Node Relation Prediction (NRP)}
This task is a pixel-only adaptation of the Node Relation Prediction objective introduced by MarkupLM\cite{li2021markuplm}, which takes the tree-structure of HTML and labels the relationship as either \{self, parent, child, sibling, ancestor, descendent, others\}. Given two elements outlined with bounding boxes in the input image, the model has to predict their node-level relationship. This task is expected to force the model to learn the relationships between the various layout components and how they interact.
\subsubsection{Table Detection}
To closely mimic the downstream task for table detection, we construct table detection on our screenshot data. The construction is as simple as merging the bounding box for the elements with \texttt{<table[id]>} contained in their Xpaths, which results in the ground truth bounding box for each table. We then ask the model to predict the following sequence:\texttt{\textcolor{brown}{<x$_{table0}$><y$_{table0}$><x$_{table0}$><y$_{table0}$><x$_{table1}$> <y$_{table1}$><x$_{table1}$><y$_{table1}$>}...}
\subsubsection{Table Parsing}
The original Screen Parsing objective, although encouraging structure-level understanding, does not emphasize the semantics of those structures, as the pre-processing replaces tags with empty brackets. We argue that the information contained in the tags is also useful signal for pre-training, especially for well-structured elements like \verb|<table>|. Therefore, we design a table parsing objective which contains the original tag name as well as the text contents for tables inside a page, as shown in Figure \ref{fig:wide_graphic}. 
\subsubsection{Screenshot Titling}

\begin{figure}
\includegraphics[scale=0.5]{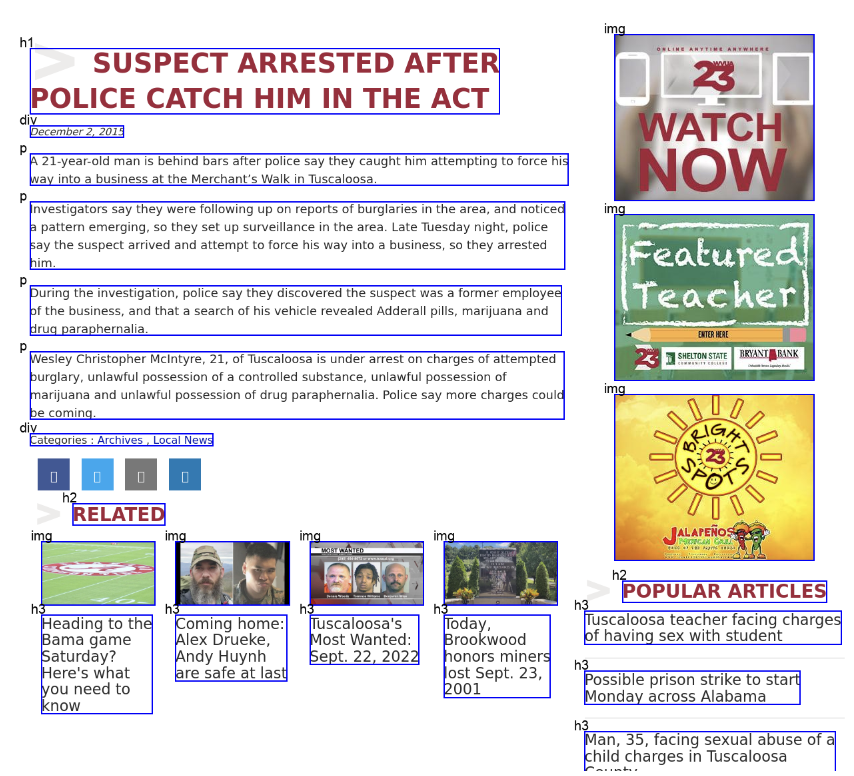}

\caption{Visualization of layout parsed from a screenshot. Corresponding HTML tags like \texttt{<h1>} are visualize on top-left corner of the bounding box.}
\vspace{-1.6em}
\label{fig:layout}
\end{figure}

To encourage the model to summarize the content in the screenshot and improve its ability on image captioning, we propose a screen titling task. Specifically, the main title in the screenshot is masked and the model is asked to generate the title text by only looking at the rest of the web page. The ground truth title is obtained from the $<title>$ node of the HTML DOM tree. The Screenshot Titling task closely resembles the screen summarization task for UI understanding, 

\subsubsection{Layout Analysis}
Obtaining layout from a screenshot is realized by grouping elements under the same sub-tree in the HTML. Specifically, for each element we obtain its \textit{cleaned Xpath} by only keeping tags in \texttt{[<p>,<table>,<form>,<dl>,<button>,<ol>,
<ul>,<nav>,<img>,<object>]} as they represent the semantic abstraction of the element. Then, we group each elements according to the value of their \textit{cleaned Xpath} to form layout of the screenshot. A visualization of the layout from a screenshot is shown in Figure \ref{fig:layout}.

\subsection{Architecture}
We adopt a simple architecture with an image encoder followed by a text decoder, same as Pix2Struct \cite{lee2023pix2struct} and similar to Donut \cite{kim2022ocrfree}. The Image encoder is ViT \cite{dosovitskiy2020image} and text decoder is transformer decoder, where the vocabulary is extended with 1000 coordinate tokens (representing discrete positions in images, normalized between 0-1000) to support localization tasks such as object detection and visual grounding. Such image-encoder-text-decoder models don't need text input and have the advantage of being OCR-free, which leads to reduced latency \cite{kim2022ocrfree}. On the other hand, in order to read textual content that are typically small, input image resolution has to be high for good performance, which leads to increased memory usage. Our proposed S4 pre-training paradigm is not limited to this architecture and can be applied to other approaches as well.

\setlength{\tabcolsep}{2pt}
\begin{table*}
\vspace{-1em}
\small
\centering

\begin{tabular}{l|ccc|ccccc}
\toprule
   \multirow{2}{*}{\bf Methods} & \bf Pre-training & \bf Pre-training& \bf Finetune & \multirow{2}{*}{\bf ChartQA $\uparrow$ } & \bf \multirow{2}{*}{\refcls $\uparrow$} & \bf Widget $\uparrow$ & \bf Screen $\uparrow$ &  \multirow{2}{*}{\bf WebSRC $\uparrow$} \\
   &\bf Dataset & \bf Objectives & \bf Batchsize &  &  & \bf Cap. & \bf Sum.\\
% \hline
\midrule

 \textcolor{gray}{Pix2Struct \cite{lee2023pix2struct}} & \textcolor{gray}{Google Priv. Data - 80M} & \textcolor{gray}{Screen Parsing}& \textcolor{gray}{32 to 256} & \textcolor{gray}{56.0} & \textcolor{gray}{92.2} & \textcolor{gray}{133.1} & \textcolor{gray}{107.0} & - \\
% \hline

\textcolor{gray}{Pix2Struct}\textsuperscript{$\dagger$} & \textcolor{gray}{Google Priv. Data - 80M} & \textcolor{gray}{Screen Parsing} & \textcolor{gray}{8} & \textcolor{gray}{54.3} & \textcolor{gray}{91.7} & \textcolor{gray}{131.1} & \textcolor{gray}{105.5} & \textcolor{gray}{60.4} \\
% \hline
\midrule

% \hline
Donut \cite{kim2022ocrfree} & SynthDoG - 37M & OCR & 64 & 41.8  & - & 127.4 & 56.4 & - \\
Pix2Struct\textsuperscript{*}& S4 Data - 2M & Screen Parsing & 8  & 47.4 & 87.9 & 129.5  & 101.3 & 58.7 \\

Pix2Struct\textsuperscript{*}& S4 Data - 15M & Screen Parsing & 8  & 52.1 & 88.1 & 129.2 & 104.5 & 60.1 \\

S4\textsuperscript{*}  \textbf{(Ours)} & S4 Data - 2M & S4$_{NL}$ & 8 & 50.5 & 92.4 & 130.5 & 103.2 & 60.5 \\
% \hline
S4\textsuperscript{*}  \textbf{(Ours)}& S4 Data - 15M & S4$_{NL}$ & 8 & \textbf{55.0} &  \textbf{94.9} & \textbf{130.6} & \textbf{105.7} & \textbf{61.1} \\

\bottomrule

\end{tabular}
\caption{Results for Chart, Web, and UI Understanding datasets. \textsuperscript{*} denotes that we load Pix2Struct's pre-trained weight and further pre-train on our S4 dataset with corresponding objectives. \textsuperscript{$\dagger$} denotes the reproduced results on downstream tasks with Pix2Struct-base's pre-trained weight and smaller batch size. Results from \textcolor{gray}{gray} rows are not directly comparable to our S4 model since we don't have access to their non-released pre-training datasets. The results from last 4 rows show that in addition to the Pix2Struct's pre-training objective, our supervised pre-training extracted from HTML DOM tree brings consistent improvement on various of downstream tasks. Note that the OCR objective for donut doesn't include bounding box prediction. \label{tab:exp_table}}
\vspace{-1em}
\end{table*}

% \newpage
\section{Experiments}
We validate the effectiveness of our ten proposed pre-training tasks by fine-tuning the model on nine downstream tasks 
and compare its performance to a Pix2Struct baseline model that was only pre-trained with screen parsing. Based on the output format, we also divide the downstream tasks into two groups.

\subsection{Implementation Details}
{\bf \noindent Pre-training schema.} We propose 2 pre-training schemes, S4$_{NL}$ for natural language generation and S4$_{Loc}$ for localization, targeting on different downstream tasks. Specifically, S4$_{NL}$ includes the baseline screen parsing task and all the tasks on natural language generation, including Attribute Prediction, Table Parsing, Title Generation, and Node Relation Prediction. S4$_{Loc}$ comprises of tasks with bounding box generations, including OCR, Image Grounding, Element Grounding, Table Detection and Layout Analysis, in addition to the screen parsing task. During pre-training, we randomly sample one task for each image with uniform distribution.

\subsubsection{Pretraining Settings}
We conducted pretraining on both 2 million and 15 million subsets of our S4 dataset, during which we set the screenshot's viewport to 1280*1280. We initialize our model with weights from Pix2struct-base and set our batch size to 32 for each node and use 4 A100 nodes during pretraining. The maximum sequence length for the pretraining targets is 128 and the patch size for the input image is 2048. Our optimizer is AdamW with learning rate set to 1e-4 with cosine decay, and for both 2M and 15M subsets we pretrain with 1 epoch per pretraining task. For instance, for S4$_{NL}$ pretraining with the 2M subset, there are 5 tasks so the total training sample is 5*2M = 10M. Note that since for each screenshots we can obtain multiple tasks, the models sees the same subset of screenshots regardless of the number of pretraining tasks.

\subsection{Chart, Web, and UI Understanding}

In this section we evaluate on tasks that require generating natural language responses from image inputs. We focus on Chart \& Web VQA, UI summarization and UI widget captioning.
\setlength{\tabcolsep}{2pt}
\begin{table*}
\centering
\small
% \resizebox{0.9\linewidth}{!}{
\begin{tabular}{l|cc|cccc}
\toprule
   \multirow{3}{*}{\bf Methods} &  \multirow{3}{*}{\begin{tabular}{@{}c@{}} \bf Pre-training \\ \bf Dataset\end{tabular}}   & \multirow{3}{*}{\begin{tabular}{@{}c@{}} \bf Pre-training \\ \bf Objectives\end{tabular}} & \bf RefExp $\uparrow$ & \bf PublayNet $\uparrow$ & \bf PubTables1M $\uparrow$ & \bf ICDAR 2019 $\uparrow$ \\
   % \hline
   & & &$_{\mathrm{cand\_free}}$ & & \bf (Table Det.) & \bf (Modern Subset)\\
 &&& \bf 30k samples & \bf 1M samples & \bf 400k samples & 
 \bf 600 samples \\

\midrule

\textcolor{gray}{DETR} & -  & - & - & - & \textcolor{gray}{99.5} & -  \\
\textcolor{gray}{DiT-B (Cascade RCNN)} & \textcolor{gray}{IIT-CDIP - 42M} & \textcolor{gray}{MIM}  & -    &\textcolor{gray}{95.4} & - & \textcolor{gray}{97.2}\\

\midrule
Pix2Struct \cite{lee2023pix2struct} & Google Priv. Data - 80M & Screen Parsing & 55.1  &  91.1 & 97.0 & 3.6 \\
% \hline

Pix2Struct\textsuperscript{*} & S4 Data - 2M & Screen Parsing & 52.7  & 91.0 & 97.1 & 3.3 \\
% \hline
S4\textsuperscript{*} (\textbf{Ours}) & S4 Data - 2M& S4$_{Loc}$ & 83.6  & 92.5 & 98.4 & 70.7 \\
S4\textsuperscript{*} (\textbf{Ours}) & S4 Data - 15M& S4$_{Loc}$ & \textbf{84.3}  & \textbf{93.1 }& \textbf{99.0} &  \textbf{79.4} \\

\bottomrule
\end{tabular}
% }
\caption{Results on Detection and Grounding datasets. We train all of the models with batch size = 32 and patch size = 2048.\textsuperscript{*} denotes that we load Pix2Struct-base's pre-trained weight and further pre-train on our S4 dataset with corresponding objectives. Models denoted \textcolor{gray}{gray} are specialist detection models that cannot parse language input (i.e cannot do grounding tasks). With our pre-training objectives, autogressive models can get significant boosts on various detection \& grounding tasks. MIM refers to Masked Image Modeling.\label{tab:exp_grounding}}
\end{table*}

\subsubsection{Datasets}

{\bf \noindent ChartQA}: ChartQA\cite{masry2022chartqa} is a VQA dataset for different types of charts (bar charts, line graphs, etc.). It includes both extractive and reasoning questions, which requires analyzing the visual data in charts to extract the relevant information. We follow the convention and report the Relaxed Match metric on ChartQA.

{\bf \noindent WebSRC}: WebSRC\cite{chen2021websrc} is a web-based VQA dataset. It contains both cleaned HTML and the screenshot of web pages, and the task is to answer the question about the content in the web page. Prior arts mostly tackle this problem by taking the ground truth cleaned HTML code as inputs, which is an unrealistic setting as real-word applications often have much more complex HTML codes than the cleaned data. Instead, our model only takes the screenshot as inputs and predicts the answer from pure-vision information. On WebSRC the Exact Match metric is reported.

{\bf \noindent Screen2words}: Screen2words\cite{wang2021screen2words} is a dataset for extracting summarization from screenshots of mobile app screens. We use Bleu and Cider scores as the evaluation metrics.

{\bf \noindent Widget Captioning}: Widget Captioning\cite{li2020widget} is a dataset for generating descriptive captions for UI widgets. The task is to generate captions that accurate describe the purpose or function of a widget in a bounding box, such as a button, slider, or a checkbox. In the input screenshot, the target widget is specified through the rendered bounding box. Bleu and Cider scores are used as evaluation metrics.

{\bf \noindent UI \refcls} UI Referential Expression (RefExp) \cite{bai2021uibert} is a dataset specifically designed for grounding referring expressions for UI elements in screenshots.
Current SOTA usually approach this problem in a simplified classification formulation: given a question and a candidate widget from annotation, the model is asked to predict whether the widget and question are related. This setting requires little localization ability from the model as the candidates widget bounding boxes are provided as inputs. We call this classification setting \refcls and report the classification accuracy as the metric.

\vspace{-2mm}
\subsubsection{Settings}

Following the same training and evaluation protocol \cite{lee2023pix2struct}, our model was pre-trained with S4$_{NL}$ objectives and finetuned on the Chart, Web, and UI Understanding tasks. Since there's no access to Google's 80M private data, we compare to two PixStruct variations. The first one is with the weights pre-trained on its private data using screen parsing released by the original author. The second one is initialized with the former's weights, and is further pre-trained on our 2M and 15M S4 data with only screen parsing objective. To have a fair comparison, our model is initialized with the same weights, and pre-trained on the same amount of data (2M and 15M S4 data) but with extra tasks. We also compare to Donut\cite{kim2022ocrfree}, which is another model uses pure-vision inputs and produces text predictions.

\subsubsection{Results}
\vspace{-2mm}

We tabulate the results in \cref{tab:exp_table}. Our method consistently outperforms Pix2Struct on all downstream tasks with significant margins when pre-trained with the same data. Specifically, when pre-trained with 15M image-text pairs, our method achieves 2.9, 6.8, 1.4, 1.2, and 1.0 improvement over Pix2Struct on ChartQA, \refcls, Widget Captioning, Screen Summarization, and WebSRC, respectively. Notice that the largest improvement is obtained on \refcls, because the pre-training tasks we proposed, such the attribute prediction and node relation prediction, help the model build robust connections between the UI elements and their referring expressions. In addition, when more data (15M) is available, our pre-training scheme S4$_{NL}$ gets improved accuracy compared to less training data (2M). The outstanding results comparing to the baseline and Donut demonstrate the efficacy of the proposed S4 pre-training scheme for downstream tasks that involve chart, web, and UI understanding.

We also noticed that, comparing to the original Pix2struct, further pre-training it on our 2M data with screen parsing objective harms the performance across different datasets. This is expected as our data collected from Common Crawl has different distribution against the original Pix2struct data due to the data size (2M vs. 80M) and potentially different website filtering strategies. Specifically, we filter out website whose CSS and JS files failed to download, while the filtering process remain unclear for Pix2struct. In addition, we also used a much smaller batch size (128 vs. 2048) due to computational constraints. Therefore, the pre-train on 2M data might drive the model weight to a less optimal state. With 15M pre-training data, we observed performance improvements over 2M data, implying that more data might compensate this distribution shift.

\subsection{Detection and Grounding}
We further investigate the effect of S4 pre-training on tasks that require spatial information understanding, such as image grounding and localization. While current SOTA models adopt specific architectures for detection, we show that with sufficient pre-training tasks on localization, an auto-gressive model can close the gap towards detection-specific architectures.

\begin{table*}[th]
\setlength{\tabcolsep}{3pt}
\centering
\scalebox{0.86}{
\begin{tabular}{c c c c c|c|c|c|c}
\toprule
  \multicolumn{1}{c|}{Titling} & \multicolumn{1}{c|}{Attibute Pred.} & \multicolumn{1}{c|}{NRP} & \multicolumn{1}{c|}{Table Parsing.} & Screen Parsing &ChartQA& Widget Cap.& Screen Sum.& \refcls \\
\midrule
& & & & \checkmark & 47.4 & 129.5 & 101.3 & 87.9 \\

& & & \checkmark & \checkmark & 48.7 & 128.3 & 101.4 & 87.8\\
& & \checkmark & \checkmark & \checkmark & \textbf{50.7} & 128.3 & 101.3 & 89.4 \\
& \checkmark & \checkmark & \checkmark & \checkmark & 50.1 & \textbf{130.7}  & 101.1 & \textbf{92.7}\\
\checkmark & \checkmark & \checkmark & \checkmark & \checkmark & 50.5 & 130.5 & \textbf{103.2} & 92.4 \\

\bottomrule
\end{tabular}
}

\vspace{0.2em}

\scalebox{0.9}{
\begin{tabular}{c c c c c|c|c}
\toprule
 \multicolumn{1}{c|}{Table Detection} & \multicolumn{1}{c|}{Layout Analysis} & \multicolumn{1}{c|}{Image \& Element Grounding } & \multicolumn{1}{c|}{OCR} & Screen Parsing & ICDAR  & \refcandf  \\
\midrule

&& & & \checkmark & 3.3 &  52.7      \\

&& & \checkmark & \checkmark & 50.1 &   68.6    \\

& &\checkmark & \checkmark & \checkmark & 52.9 &   66.2  \\

\checkmark &\checkmark & \checkmark & \checkmark & \checkmark & \textbf{70.7} & \textbf{83.6}     \\

\bottomrule
\end{tabular}
}
\caption{Ablation study on adding different pre-training objectives using 2M S4 data.  \label{tab:task_ablation}}
\vspace{-1em}
\end{table*}

\subsubsection{Datasets}

\noindent \textbf{PubLayNet}: PubLayNet \cite{zhong2019publaynet} is a large-scale dataset for document layout analysis, containing more than 360,000 pages of scientific articles. We evaluate on the bounding box prediction task and report AP 50 as the metric.
    
\noindent \textbf{ICDAR2019}: ICDAR2019 \cite{8978120} is a table detection dataset that contains 1200 modern and archived documents. We only evaluate on the modern document split as the archived documents do not have bounding boxes. We use AP 50 as our evaluation metric.

\noindent \textbf{PubTables-1M} PubTables-1M \cite{smock2022pubtables} has around 400k images with 947,642 tables from PMCOA scientific articles and we use it for table detection experiments. AP 50 is used as the evaluation metric.

\noindent \textbf{UI \refcandf}
As mentioned earlier, current works mostly treat the UI RefExp task as a binary classification problem using the ground truth bounding boxes as candidates, making it less challenging as it does not measure whether the model can localize the UI elements. In this work, we propose a new task, \textbf{UI \refcandf}, to include the element grounding into the challenges. In this task, the input is only the screenshot with the text description, and the model is asked to predict the bounding box of the related UI element directly, thus "candidate free". For evaluation, the predicted bounding box will be matched to the closest ground truth box to compute the accuracy.

\subsubsection{Settings}

Our model was pre-trained with S4${_{Loc}}$ for the benefits on localization related tasks. The model is then fine-tuned and evaluated on each downstream task dataset. We compare to two PixStruct variations. The first one is the weights pre-trained on its private data using screen parsing released by the original author. The second one is initialized with the former weights and is further pre-trained on our S4 data 2M with only screen parsing, to be compared to our model pre-trained on the same amount of data.

\subsubsection{Results}

The evaluation results on all detection and grounding related tasks are tabulated in \cref{tab:exp_grounding}. Our method shows clear advantages over the baseline Pix2Struct model that was only pre-trained with the screen parsing task. Specifically, on \refcandf, when pre-trained with 2M data, our method outperforms Pix2Struct by a significant margin of 30.9 (83.6 vs. 52.7). This is because our pre-training tasks like OCR prediction and element grounding, help the model learn to localize items in the image given their semantic descriptions. Similarly, on ICDAR, which only has 600 training samples, our method achieves 70.7 AP with 2M pre-training data, while the baseline Pix2Struct only obtains 3.3, due to its lack of localization ability. When there are more fine-tuning data for the downstream tasks (PublayNet 1M \& PubTables 400K), Pix2Struct can learn to localize the objects and obtain decent performance, but our training scheme S4$_{Loc}$ still benefits the model and improves the performance by 1.5 on PublayNet and 1.3 on PubTables. 

The benefits of S4$_{Loc}$ pre-training becomes more prominent when more pre-train data is available. Our model pre-trained with 15M data consistently improves the accuracy on all four downstream tasks compared to 2M pre-train data. In particular, on ICDAR the 15M pre-train data improves the accuracy from 70.7 to 79.4, showing that the pre-training task benefits more when the downstream task has less data. It is worth noting that, as a generic auto-regressive text generation model, our method with 15M pre-training data achieves comparable performance on PublayNet and PubTables to detection specific models like DeTR and Dit-B, showing that sufficient pre-training with proper tasks helps close the gap between auto-regressive models and detection-specific architectures.

\subsection{Contribution of each task}

We conducted ablative studies to show the effectiveness of each individual pre-training tasks besides screen parsing. For S4$_{NL}$, we evaluate on ChartQA, Widget Captioning, Screen Summarization, and \refcls, by adding the natural language related tasks gradually. For S4$_{Loc}$, we also add the localization related tasks incrementally and evaluate on \refcandf~and ICDAR. The results are shown in \cref{tab:task_ablation}. Observe that the downstream task usually benefits from the addition of the most related pre-training task. For example, Screen Summarization gets 2.1 performance improvement when the screen tilting pre-training task is added, while the other tasks have little effect on the performance. The attribute prediction task encourages the model to associate website elements to their text description. Therefore, adding it to the pre-training scheme significantly improves the performance on both Widget Captioning and \refcls, which requires the model to associate UI elements to texts. Similarly, adding all the localization related pre-train task substantially improves the model's ability on grounding elements, resulting in higher accuracy on both ICDAR and \refcandf.

\section{Conclusions}
We introduced a novel pre-training framework for vision-language models, with which models are exposed with a variety of supervised tasks on diverse and massive amount website screenshots. This innovation is enabled by our proposed data pipeline, in which the web pages are rendered, extracted, and cleaned automatically to generate screenshots and corresponding annotations. The tasks in our pre-training scheme are designed to maximize the utilization of annotations in our data as well as the similarities between the downstream tasks. Through extensive experiments, we demonstrated the efficacy of our method on boosting the downstream tasks performance on 9 different datasets.

\newpage
{
    \small
    \bibliographystyle{ieeenat_fullname}
    \bibliography{main}

\begin{thebibliography}{76}
\providecommand{\natexlab}[1]{#1}
\providecommand{\url}[1]{\texttt{#1}}
\expandafter\ifx\csname urlstyle\endcsname\relax
  \providecommand{\doi}[1]{doi: #1}\else
  \providecommand{\doi}{doi: \begingroup \urlstyle{rm}\Url}\fi

\bibitem[Appalaraju et~al.(2021)Appalaraju, Jasani, Kota, Xie, and Manmatha]{Appalaraju_2021_ICCV}
Srikar Appalaraju, Bhavan Jasani, Bhargava~Urala Kota, Yusheng Xie, and R. Manmatha.
\newblock Docformer: End-to-end transformer for document understanding.
\newblock In \emph{Proceedings of the IEEE/CVF International Conference on Computer Vision (ICCV)}, pages 993--1003, 2021.

\bibitem[Appalaraju et~al.(2024)Appalaraju, Tang, Dong, Sankaran, Zhou, and Manmatha]{Appalaraju2023DocFormerv2LFAAAI}
Srikar Appalaraju, Peng Tang, Qi Dong, Nishant Sankaran, Yichu Zhou, and R. Manmatha.
\newblock Docformerv2: Local features for document understanding.
\newblock \emph{AAAI}, abs/2306.01733, 2024.

\bibitem[Arici et~al.(2021)Arici, Seyfioglu, Neiman, Xu, Train, Chilimbi, Zeng, and Tutar]{arici2021mlim}
Tarik Arici, Mehmet~Saygin Seyfioglu, Tal Neiman, Yi Xu, Son Train, Trishul Chilimbi, Belinda Zeng, and Ismail Tutar.
\newblock Mlim: Vision-and-language model pre-training with masked language and image modeling.
\newblock \emph{arXiv preprint arXiv:2109.12178}, 2021.

\bibitem[Bai et~al.(2021)Bai, Zang, Xu, Sunkara, Rastogi, Chen, et~al.]{bai2021uibert}
Chongyang Bai, Xiaoxue Zang, Ying Xu, Srinivas Sunkara, Abhinav Rastogi, Jindong Chen, et~al.
\newblock Uibert: Learning generic multimodal representations for ui understanding.
\newblock \emph{arXiv preprint arXiv:2107.13731}, 2021.

\bibitem[Bao et~al.(2021)Bao, Dong, Piao, and Wei]{bao2021beit}
Hangbo Bao, Li Dong, Songhao Piao, and Furu Wei.
\newblock Beit: Bert pre-training of image transformers.
\newblock \emph{arXiv preprint arXiv:2106.08254}, 2021.

\bibitem[Biten et~al.(2022)Biten, Litman, Xie, Appalaraju, and Manmatha]{Biten_2022_CVPR}
Ali~Furkan Biten, Ron Litman, Yusheng Xie, Srikar Appalaraju, and R. Manmatha.
\newblock Latr: Layout-aware transformer for scene-text vqa.
\newblock In \emph{Proceedings of the IEEE/CVF Conference on Computer Vision and Pattern Recognition (CVPR)}, pages 16548--16558, 2022.

\bibitem[Brown et~al.(2020)Brown, Mann, Ryder, Subbiah, Kaplan, Dhariwal, Neelakantan, Shyam, Sastry, Askell, et~al.]{brown2020language}
Tom Brown, Benjamin Mann, Nick Ryder, Melanie Subbiah, Jared~D Kaplan, Prafulla Dhariwal, Arvind Neelakantan, Pranav Shyam, Girish Sastry, Amanda Askell, et~al.
\newblock Language models are few-shot learners.
\newblock \emph{Advances in neural information processing systems}, 33:\penalty0 1877--1901, 2020.

\bibitem[Chen et~al.(2022{\natexlab{a}})Chen, Zhang, Han, Chen, Shi, Xu, and Xu]{Chen2022VLPAS}
Feilong Chen, Duzhen Zhang, Minglun Han, Xiuyi Chen, Jing Shi, Shuang Xu, and Bo Xu.
\newblock Vlp: A survey on vision-language pre-training.
\newblock \emph{Machine Intelligence Research}, 20:\penalty0 38--56, 2022{\natexlab{a}}.

\bibitem[Chen et~al.(2021)Chen, Zhao, Chen, Zhang, Ji, Luo, Xiong, and Yu]{chen2021websrc}
Xingyu Chen, Zihan Zhao, Lu Chen, Danyang Zhang, Jiabao Ji, Ao Luo, Yuxuan Xiong, and Kai Yu.
\newblock Websrc: A dataset for web-based structural reading comprehension, 2021.

\bibitem[Chen et~al.(2022{\natexlab{b}})Chen, Wang, Changpinyo, Piergiovanni, Padlewski, Salz, Goodman, Grycner, Mustafa, Beyer, et~al.]{chen2022pali}
Xi Chen, Xiao Wang, Soravit Changpinyo, AJ Piergiovanni, Piotr Padlewski, Daniel Salz, Sebastian Goodman, Adam Grycner, Basil Mustafa, Lucas Beyer, et~al.
\newblock Pali: A jointly-scaled multilingual language-image model.
\newblock \emph{arXiv preprint arXiv:2209.06794}, 2022{\natexlab{b}}.

\bibitem[Chung et~al.(2022)Chung, Hou, Longpre, Zoph, Tay, Fedus, Li, Wang, Dehghani, Brahma, et~al.]{chung2022scaling}
Hyung~Won Chung, Le Hou, Shayne Longpre, Barret Zoph, Yi Tay, William Fedus, Yunxuan Li, Xuezhi Wang, Mostafa Dehghani, Siddhartha Brahma, et~al.
\newblock Scaling instruction-finetuned language models.
\newblock \emph{arXiv preprint arXiv:2210.11416}, 2022.

\bibitem[Deng et~al.(2009)Deng, Dong, Socher, Li, Li, and Fei-Fei]{deng2009imagenet}
Jia Deng, Wei Dong, Richard Socher, Li-Jia Li, Kai Li, and Li Fei-Fei.
\newblock Imagenet: A large-scale hierarchical image database.
\newblock In \emph{2009 IEEE conference on computer vision and pattern recognition}, pages 248--255. Ieee, 2009.

\bibitem[Devlin et~al.(2018)Devlin, Chang, Lee, and Toutanova]{devlin2018bert}
Jacob Devlin, Ming-Wei Chang, Kenton Lee, and Kristina Toutanova.
\newblock Bert: Pre-training of deep bidirectional transformers for language understanding.
\newblock \emph{arXiv preprint arXiv:1810.04805}, 2018.

\bibitem[Dosovitskiy et~al.(2020)Dosovitskiy, Beyer, Kolesnikov, Weissenborn, Zhai, Unterthiner, Dehghani, Minderer, Heigold, Gelly, et~al.]{dosovitskiy2020image}
Alexey Dosovitskiy, Lucas Beyer, Alexander Kolesnikov, Dirk Weissenborn, Xiaohua Zhai, Thomas Unterthiner, Mostafa Dehghani, Matthias Minderer, Georg Heigold, Sylvain Gelly, et~al.
\newblock An image is worth 16x16 words: Transformers for image recognition at scale.
\newblock \emph{arXiv preprint arXiv:2010.11929}, 2020.

\bibitem[Dou et~al.(2022)Dou, Kamath, Gan, Zhang, Wang, Li, Liu, Liu, LeCun, Peng, Gao, and Wang]{Dou2022CoarsetoFineVP}
Zi-Yi Dou, Aishwarya Kamath, Zhe Gan, Pengchuan Zhang, Jianfeng Wang, Linjie Li, Zicheng Liu, Ce Liu, Yann LeCun, Nanyun Peng, Jianfeng Gao, and Lijuan Wang.
\newblock Coarse-to-fine vision-language pre-training with fusion in the backbone.
\newblock \emph{ArXiv}, abs/2206.07643, 2022.

\bibitem[Du et~al.(2022)Du, Liu, Li, and Zhao]{Du2022ASO}
Yifan Du, Zikang Liu, Junyi Li, and Wayne~Xin Zhao.
\newblock A survey of vision-language pre-trained models.
\newblock In \emph{International Joint Conference on Artificial Intelligence}, 2022.

\bibitem[Gao et~al.(2019)Gao, Huang, Déjean, Meunier, Yan, Fang, Kleber, and Lang]{8978120}
Liangcai Gao, Yilun Huang, Hervé Déjean, Jean-Luc Meunier, Qinqin Yan, Yu Fang, Florian Kleber, and Eva Lang.
\newblock Icdar 2019 competition on table detection and recognition (ctdar).
\newblock In \emph{2019 International Conference on Document Analysis and Recognition (ICDAR)}, pages 1510--1515, 2019.

\bibitem[Gao et~al.(2020)Gao, Biderman, Black, Golding, Hoppe, Foster, Phang, He, Thite, Nabeshima, Presser, and Leahy]{pile}
Leo Gao, Stella Biderman, Sid Black, Laurence Golding, Travis Hoppe, Charles Foster, Jason Phang, Horace He, Anish Thite, Noa Nabeshima, Shawn Presser, and Connor Leahy.
\newblock The {P}ile: An 800gb dataset of diverse text for language modeling.
\newblock \emph{arXiv preprint arXiv:2101.00027}, 2020.

\bibitem[Gao et~al.(2021)Gao, Geng, Zhang, Ma, Fang, Zhang, Li, and Qiao]{Gao2021CLIPAdapterBV}
Peng Gao, Shijie Geng, Renrui Zhang, Teli Ma, Rongyao Fang, Yongfeng Zhang, Hongsheng Li, and Yu~Jiao Qiao.
\newblock Clip-adapter: Better vision-language models with feature adapters.
\newblock \emph{ArXiv}, abs/2110.04544, 2021.

\bibitem[Goyal et~al.(2017)Goyal, Khot, Summers-Stay, Batra, and Parikh]{goyal2017making}
Yash Goyal, Tejas Khot, Douglas Summers-Stay, Dhruv Batra, and Devi Parikh.
\newblock Making the v in vqa matter: Elevating the role of image understanding in visual question answering.
\newblock In \emph{Proceedings of the IEEE conference on computer vision and pattern recognition}, pages 6904--6913, 2017.

\bibitem[He et~al.(2022)He, Chen, Xie, Li, Doll{\'a}r, and Girshick]{he2022masked}
Kaiming He, Xinlei Chen, Saining Xie, Yanghao Li, Piotr Doll{\'a}r, and Ross Girshick.
\newblock Masked autoencoders are scalable vision learners.
\newblock In \emph{Proceedings of the IEEE/CVF conference on computer vision and pattern recognition}, pages 16000--16009, 2022.

\bibitem[Hu et~al.(2021)Hu, Gan, Wang, Yang, Liu, Lu, and Wang]{Hu2021ScalingUV}
Xiaowei Hu, Zhe Gan, Jianfeng Wang, Zhengyuan Yang, Zicheng Liu, Yumao Lu, and Lijuan Wang.
\newblock Scaling up vision-language pretraining for image captioning.
\newblock \emph{2022 IEEE/CVF Conference on Computer Vision and Pattern Recognition (CVPR)}, pages 17959--17968, 2021.

\bibitem[Huang et~al.(2021)Huang, Zeng, Huang, Liu, Fu, and Fu]{Huang2021SeeingOO}
Zhicheng Huang, Zhaoyang Zeng, Yupan Huang, Bei Liu, Dongmei Fu, and Jianlong Fu.
\newblock Seeing out of the box: End-to-end pre-training for vision-language representation learning.
\newblock \emph{2021 IEEE/CVF Conference on Computer Vision and Pattern Recognition (CVPR)}, pages 12971--12980, 2021.

\bibitem[Jia et~al.(2021)Jia, Yang, Xia, Chen, Parekh, Pham, Le, Sung, Li, and Duerig]{Jia2021ScalingUV}
Chao Jia, Yinfei Yang, Ye Xia, Yi-Ting Chen, Zarana Parekh, Hieu Pham, Quoc~V. Le, Yun-Hsuan Sung, Zhen Li, and Tom Duerig.
\newblock Scaling up visual and vision-language representation learning with noisy text supervision.
\newblock In \emph{International Conference on Machine Learning}, 2021.

\bibitem[Jin et~al.(2021)Jin, Cheng, Shen, Chen, and Ren]{Jin2021AGP}
Woojeong Jin, Yu Cheng, Yelong Shen, Weizhu Chen, and Xiang Ren.
\newblock A good prompt is worth millions of parameters: Low-resource prompt-based learning for vision-language models.
\newblock In \emph{Annual Meeting of the Association for Computational Linguistics}, 2021.

\bibitem[Kamath et~al.(2021)Kamath, Singh, LeCun, Misra, Synnaeve, and Carion]{Kamath2021MDETRM}
Aishwarya Kamath, Mannat Singh, Yann LeCun, Ishan Misra, Gabriel Synnaeve, and Nicolas Carion.
\newblock Mdetr - modulated detection for end-to-end multi-modal understanding.
\newblock \emph{2021 IEEE/CVF International Conference on Computer Vision (ICCV)}, pages 1760--1770, 2021.

\bibitem[Kim et~al.(2022)Kim, Hong, Yim, Nam, Park, Yim, Hwang, Yun, Han, and Park]{kim2022ocrfree}
Geewook Kim, Teakgyu Hong, Moonbin Yim, Jeongyeon Nam, Jinyoung Park, Jinyeong Yim, Wonseok Hwang, Sangdoo Yun, Dongyoon Han, and Seunghyun Park.
\newblock Ocr-free document understanding transformer, 2022.

\bibitem[Kirillov et~al.(2023)Kirillov, Mintun, Ravi, Mao, Rolland, Gustafson, Xiao, Whitehead, Berg, Lo, et~al.]{kirillov2023segment}
Alexander Kirillov, Eric Mintun, Nikhila Ravi, Hanzi Mao, Chloe Rolland, Laura Gustafson, Tete Xiao, Spencer Whitehead, Alexander~C Berg, Wan-Yen Lo, et~al.
\newblock Segment anything.
\newblock \emph{arXiv preprint arXiv:2304.02643}, 2023.

\bibitem[Kuznetsova et~al.(2020)Kuznetsova, Rom, Alldrin, Uijlings, Krasin, Pont-Tuset, Kamali, Popov, Malloci, Kolesnikov, et~al.]{kuznetsova2020open}
Alina Kuznetsova, Hassan Rom, Neil Alldrin, Jasper Uijlings, Ivan Krasin, Jordi Pont-Tuset, Shahab Kamali, Stefan Popov, Matteo Malloci, Alexander Kolesnikov, et~al.
\newblock The open images dataset v4: Unified image classification, object detection, and visual relationship detection at scale.
\newblock \emph{International Journal of Computer Vision}, 128\penalty0 (7):\penalty0 1956--1981, 2020.

\bibitem[Kwon et~al.(2022)Kwon, Cai, Ravichandran, Bas, Bhotika, and Soatto]{kwon2022masked}
Gukyeong Kwon, Zhaowei Cai, Avinash Ravichandran, Erhan Bas, Rahul Bhotika, and Stefano Soatto.
\newblock Masked vision and language modeling for multi-modal representation learning.
\newblock \emph{arXiv preprint arXiv:2208.02131}, 2022.

\bibitem[Lazarow et~al.(2020)Lazarow, Lee, Shi, and Tu]{lazarow2020learning}
Justin Lazarow, Kwonjoon Lee, Kunyu Shi, and Zhuowen Tu.
\newblock Learning instance occlusion for panoptic segmentation.
\newblock In \emph{Proceedings of the IEEE/CVF conference on computer vision and pattern recognition}, pages 10720--10729, 2020.

\bibitem[Lee et~al.(2023)Lee, Joshi, Turc, Hu, Liu, Eisenschlos, Khandelwal, Shaw, Chang, and Toutanova]{lee2023pix2struct}
Kenton Lee, Mandar Joshi, Iulia Turc, Hexiang Hu, Fangyu Liu, Julian Eisenschlos, Urvashi Khandelwal, Peter Shaw, Ming-Wei Chang, and Kristina Toutanova.
\newblock Pix2struct: Screenshot parsing as pretraining for visual language understanding, 2023.

\bibitem[Li and Li(2023)]{li2023spotlight}
Gang Li and Yang Li.
\newblock Spotlight: Mobile ui understanding using vision-language models with a focus.
\newblock 2023.

\bibitem[Li et~al.(2021{\natexlab{a}})Li, Xu, Cui, and Wei]{li2021markuplm}
Junlong Li, Yiheng Xu, Lei Cui, and Furu Wei.
\newblock Markuplm: Pre-training of text and markup language for visually-rich document understanding.
\newblock 2021{\natexlab{a}}.

\bibitem[Li et~al.(2022)Li, Li, Xiong, and Hoi]{Li2022BLIPBL}
Junnan Li, Dongxu Li, Caiming Xiong, and Steven C.~H. Hoi.
\newblock Blip: Bootstrapping language-image pre-training for unified vision-language understanding and generation.
\newblock In \emph{International Conference on Machine Learning}, 2022.

\bibitem[Li et~al.(2021{\natexlab{b}})Li, Zhang, Zhang, Yang, Li, Zhong, Wang, Yuan, Zhang, Hwang, Chang, and Gao]{Li2021GroundedLP}
Liunian~Harold Li, Pengchuan Zhang, Haotian Zhang, Jianwei Yang, Chunyuan Li, Yiwu Zhong, Lijuan Wang, Lu Yuan, Lei Zhang, Jenq-Neng Hwang, Kai-Wei Chang, and Jianfeng Gao.
\newblock Grounded language-image pre-training.
\newblock \emph{2022 IEEE/CVF Conference on Computer Vision and Pattern Recognition (CVPR)}, pages 10955--10965, 2021{\natexlab{b}}.

\bibitem[Li et~al.(2020{\natexlab{a}})Li, Yin, Li, Hu, Zhang, Zhang, Wang, Hu, Dong, Wei, Choi, and Gao]{li2020oscar}
Xiujun Li, Xi Yin, Chunyuan Li, Xiaowei Hu, Pengchuan Zhang, Lei Zhang, Lijuan Wang, Houdong Hu, Li Dong, Furu Wei, Yejin Choi, and Jianfeng Gao.
\newblock Oscar: Object-semantics aligned pre-training for vision-language tasks.
\newblock \emph{ECCV 2020}, 2020{\natexlab{a}}.

\bibitem[Li et~al.(2020{\natexlab{b}})Li, Li, He, Zheng, Li, and Guan]{li2020widget}
Yang Li, Gang Li, Luheng He, Jingjie Zheng, Hong Li, and Zhiwei Guan.
\newblock Widget captioning: Generating natural language description for mobile user interface elements, 2020{\natexlab{b}}.

\bibitem[Lin et~al.(2014)Lin, Maire, Belongie, Hays, Perona, Ramanan, Doll{\'a}r, and Zitnick]{lin2014microsoft}
Tsung-Yi Lin, Michael Maire, Serge Belongie, James Hays, Pietro Perona, Deva Ramanan, Piotr Doll{\'a}r, and C~Lawrence Zitnick.
\newblock Microsoft coco: Common objects in context.
\newblock In \emph{Computer Vision--ECCV 2014: 13th European Conference, Zurich, Switzerland, September 6-12, 2014, Proceedings, Part V 13}, pages 740--755. Springer, 2014.

\bibitem[Lu et~al.(2019)Lu, Batra, Parikh, and Lee]{Lu2019ViLBERTPT}
Jiasen Lu, Dhruv Batra, Devi Parikh, and Stefan Lee.
\newblock Vilbert: Pretraining task-agnostic visiolinguistic representations for vision-and-language tasks.
\newblock In \emph{Neural Information Processing Systems}, 2019.

\bibitem[Lu et~al.(2022)Lu, Clark, Zellers, Mottaghi, and Kembhavi]{Lu2022UnifiedIOAU}
Jiasen Lu, Christopher Clark, Rowan Zellers, Roozbeh Mottaghi, and Aniruddha Kembhavi.
\newblock Unified-io: A unified model for vision, language, and multi-modal tasks.
\newblock \emph{ArXiv}, abs/2206.08916, 2022.

\bibitem[Masry et~al.(2022)Masry, Long, Tan, Joty, and Hoque]{masry2022chartqa}
Ahmed Masry, Do~Xuan Long, Jia~Qing Tan, Shafiq Joty, and Enamul Hoque.
\newblock Chartqa: A benchmark for question answering about charts with visual and logical reasoning, 2022.

\bibitem[Ouyang et~al.(2022)Ouyang, Wu, Jiang, Almeida, Wainwright, Mishkin, Zhang, Agarwal, Slama, Ray, et~al.]{ouyang2022training}
Long Ouyang, Jeffrey Wu, Xu Jiang, Diogo Almeida, Carroll Wainwright, Pamela Mishkin, Chong Zhang, Sandhini Agarwal, Katarina Slama, Alex Ray, et~al.
\newblock Training language models to follow instructions with human feedback.
\newblock \emph{Advances in Neural Information Processing Systems}, 35:\penalty0 27730--27744, 2022.

\bibitem[Radford et~al.(2018)Radford, Narasimhan, Salimans, Sutskever, et~al.]{radford2018improving}
Alec Radford, Karthik Narasimhan, Tim Salimans, Ilya Sutskever, et~al.
\newblock Improving language understanding by generative pre-training.
\newblock 2018.

\bibitem[Radford et~al.(2021{\natexlab{a}})Radford, Kim, Hallacy, Ramesh, Goh, Agarwal, Sastry, Askell, Mishkin, Clark, Krueger, and Sutskever]{Radford2021LearningTV}
Alec Radford, Jong~Wook Kim, Chris Hallacy, Aditya Ramesh, Gabriel Goh, Sandhini Agarwal, Girish Sastry, Amanda Askell, Pamela Mishkin, Jack Clark, Gretchen Krueger, and Ilya Sutskever.
\newblock Learning transferable visual models from natural language supervision.
\newblock In \emph{International Conference on Machine Learning}, 2021{\natexlab{a}}.

\bibitem[Radford et~al.(2021{\natexlab{b}})Radford, Kim, Hallacy, Ramesh, Goh, Agarwal, Sastry, Askell, Mishkin, Clark, et~al.]{radford2021learning}
Alec Radford, Jong~Wook Kim, Chris Hallacy, Aditya Ramesh, Gabriel Goh, Sandhini Agarwal, Girish Sastry, Amanda Askell, Pamela Mishkin, Jack Clark, et~al.
\newblock Learning transferable visual models from natural language supervision.
\newblock In \emph{International conference on machine learning}, pages 8748--8763. PMLR, 2021{\natexlab{b}}.

\bibitem[Raffel et~al.(2019)Raffel, Shazeer, Roberts, Lee, Narang, Matena, Zhou, Li, and Liu]{2019t5_C4dataset}
Colin Raffel, Noam Shazeer, Adam Roberts, Katherine Lee, Sharan Narang, Michael Matena, Yanqi Zhou, Wei Li, and Peter~J. Liu.
\newblock Exploring the limits of transfer learning with a unified text-to-text transformer.
\newblock \emph{arXiv e-prints}, 2019.

\bibitem[Raffel et~al.(2020)Raffel, Shazeer, Roberts, Lee, Narang, Matena, Zhou, Li, and Liu]{raffel2020exploring}
Colin Raffel, Noam Shazeer, Adam Roberts, Katherine Lee, Sharan Narang, Michael Matena, Yanqi Zhou, Wei Li, and Peter~J Liu.
\newblock Exploring the limits of transfer learning with a unified text-to-text transformer.
\newblock \emph{The Journal of Machine Learning Research}, 21\penalty0 (1):\penalty0 5485--5551, 2020.

\bibitem[Rombach et~al.(2022)Rombach, Blattmann, Lorenz, Esser, and Ommer]{rombach2022high}
Robin Rombach, Andreas Blattmann, Dominik Lorenz, Patrick Esser, and Bj{\"o}rn Ommer.
\newblock High-resolution image synthesis with latent diffusion models.
\newblock In \emph{Proceedings of the IEEE/CVF conference on computer vision and pattern recognition}, pages 10684--10695, 2022.

\bibitem[Schuhmann et~al.(2022)Schuhmann, Beaumont, Vencu, Gordon, Wightman, Cherti, Coombes, Katta, Mullis, Wortsman, et~al.]{schuhmann2022laion}
Christoph Schuhmann, Romain Beaumont, Richard Vencu, Cade Gordon, Ross Wightman, Mehdi Cherti, Theo Coombes, Aarush Katta, Clayton Mullis, Mitchell Wortsman, et~al.
\newblock Laion-5b: An open large-scale dataset for training next generation image-text models.
\newblock \emph{Advances in Neural Information Processing Systems}, 35:\penalty0 25278--25294, 2022.

\bibitem[Shao et~al.(2019)Shao, Li, Zhang, Peng, Yu, Zhang, Li, and Sun]{shao2019objects365}
Shuai Shao, Zeming Li, Tianyuan Zhang, Chao Peng, Gang Yu, Xiangyu Zhang, Jing Li, and Jian Sun.
\newblock Objects365: A large-scale, high-quality dataset for object detection.
\newblock In \emph{Proceedings of the IEEE/CVF international conference on computer vision}, pages 8430--8439, 2019.

\bibitem[Shi et~al.(2024)Shi, Dong, Goncalves, Tu, and Soatto]{Shi_2024_CVPR}
Kunyu Shi, Qi Dong, Luis Goncalves, Zhuowen Tu, and Stefano Soatto.
\newblock Non-autoregressive sequence-to-sequence vision-language models.
\newblock In \emph{Proceedings of the IEEE/CVF Conference on Computer Vision and Pattern Recognition (CVPR)}, pages 13603--13612, 2024.

\bibitem[Shu et~al.(2022)Shu, Nie, Huang, Yu, Goldstein, Anandkumar, and Xiao]{Shu2022TestTimePT}
Manli Shu, Weili Nie, De-An Huang, Zhiding Yu, Tom Goldstein, Anima Anandkumar, and Chaowei Xiao.
\newblock Test-time prompt tuning for zero-shot generalization in vision-language models.
\newblock \emph{ArXiv}, abs/2209.07511, 2022.

\bibitem[Smock et~al.(2022)Smock, Pesala, and Abraham]{smock2022pubtables}
Brandon Smock, Rohith Pesala, and Robin Abraham.
\newblock Pub{T}ables-1{M}: Towards comprehensive table extraction from unstructured documents.
\newblock In \emph{Proceedings of the IEEE/CVF Conference on Computer Vision and Pattern Recognition (CVPR)}, pages 4634--4642, 2022.

\bibitem[Wang et~al.(2021{\natexlab{a}})Wang, Li, Zhou, Chen, Grossman, and Li]{wang2021screen2words}
Bryan Wang, Gang Li, Xin Zhou, Zhourong Chen, Tovi Grossman, and Yang Li.
\newblock Screen2words: Automatic mobile ui summarization with multimodal learning, 2021{\natexlab{a}}.

\bibitem[Wang et~al.(2022{\natexlab{a}})Wang, Yang, Men, Lin, Bai, Li, Ma, Zhou, Zhou, and Yang]{Wang2022OFAUA}
Peng Wang, An Yang, Rui Men, Junyang Lin, Shuai Bai, Zhikang Li, Jianxin Ma, Chang Zhou, Jingren Zhou, and Hongxia Yang.
\newblock Ofa: Unifying architectures, tasks, and modalities through a simple sequence-to-sequence learning framework.
\newblock In \emph{International Conference on Machine Learning}, 2022{\natexlab{a}}.

\bibitem[Wang et~al.(2022{\natexlab{b}})Wang, Yang, Men, Lin, Bai, Li, Ma, Zhou, Zhou, and Yang]{wang2022ofa}
Peng Wang, An Yang, Rui Men, Junyang Lin, Shuai Bai, Zhikang Li, Jianxin Ma, Chang Zhou, Jingren Zhou, and Hongxia Yang.
\newblock Ofa: Unifying architectures, tasks, and modalities through a simple sequence-to-sequence learning framework.
\newblock In \emph{International Conference on Machine Learning}, pages 23318--23340. PMLR, 2022{\natexlab{b}}.

\bibitem[Wang et~al.(2021{\natexlab{b}})Wang, Bao, Dong, and Wei]{Wang2021VLMoUV}
Wenhui Wang, Hangbo Bao, Li Dong, and Furu Wei.
\newblock Vlmo: Unified vision-language pre-training with mixture-of-modality-experts.
\newblock \emph{ArXiv}, abs/2111.02358, 2021{\natexlab{b}}.

\bibitem[Wang et~al.(2022{\natexlab{c}})Wang, Bao, Dong, Bjorck, Peng, Liu, Aggarwal, Mohammed, Singhal, Som, and Wei]{Wang2022ImageAA}
Wenhui Wang, Hangbo Bao, Li Dong, Johan Bjorck, Zhiliang Peng, Qiang Liu, Kriti Aggarwal, Owais~Khan Mohammed, Saksham Singhal, Subhojit Som, and Furu Wei.
\newblock Image as a foreign language: Beit pretraining for all vision and vision-language tasks.
\newblock \emph{ArXiv}, abs/2208.10442, 2022{\natexlab{c}}.

\bibitem[Xie et~al.(2022)Xie, Zhang, Cao, Lin, Bao, Yao, Dai, and Hu]{xie2022simmim}
Zhenda Xie, Zheng Zhang, Yue Cao, Yutong Lin, Jianmin Bao, Zhuliang Yao, Qi Dai, and Han Hu.
\newblock Simmim: A simple framework for masked image modeling.
\newblock In \emph{Proceedings of the IEEE/CVF Conference on Computer Vision and Pattern Recognition}, pages 9653--9663, 2022.

\bibitem[Xu et~al.(2021)Xu, Yan, Li, Bi, Huang, Xiao, and Huang]{Xu2021E2EVLPEV}
Haiyang Xu, Ming Yan, Chenliang Li, Bin Bi, Songfang Huang, Wenming Xiao, and Fei Huang.
\newblock E2e-vlp: End-to-end vision-language pre-training enhanced by visual learning.
\newblock \emph{ArXiv}, abs/2106.01804, 2021.

\bibitem[Yang et~al.(2022)Yang, Duan, Tran, Xu, Chanda, Chen, Zeng, Chilimbi, and Huang]{Yang2022VisionLanguagePW}
Jinyu Yang, Jiali Duan, S. Tran, Yi Xu, Sampath Chanda, Liqun Chen, Belinda Zeng, Trishul~M. Chilimbi, and Junzhou Huang.
\newblock Vision-language pre-training with triple contrastive learning.
\newblock \emph{2022 IEEE/CVF Conference on Computer Vision and Pattern Recognition (CVPR)}, pages 15650--15659, 2022.

\bibitem[Yang et~al.(2021)Yang, Zhang, Qi, and Cai]{Yang2021CausalAF}
Xu Yang, Hanwang Zhang, Guojun Qi, and Jianfei Cai.
\newblock Causal attention for vision-language tasks.
\newblock \emph{2021 IEEE/CVF Conference on Computer Vision and Pattern Recognition (CVPR)}, pages 9842--9852, 2021.

\bibitem[Yao et~al.(2021)Yao, Zhang, Zhang, Liu, seng Chua, and Sun]{Yao2021CPTCP}
Yuan Yao, Ao Zhang, Zhengyan Zhang, Zhiyuan Liu, Tat seng Chua, and Maosong Sun.
\newblock Cpt: Colorful prompt tuning for pre-trained vision-language models.
\newblock \emph{ArXiv}, abs/2109.11797, 2021.

\bibitem[Zeng et~al.(2022)Zeng, Wong, Welker, Choromanski, Tombari, Purohit, Ryoo, Sindhwani, Lee, Vanhoucke, and Florence]{Zeng2022SocraticMC}
Andy Zeng, Adrian~S. Wong, Stefan Welker, Krzysztof Choromanski, Federico Tombari, Aveek Purohit, Michael~S. Ryoo, Vikas Sindhwani, Johnny Lee, Vincent Vanhoucke, and Peter~R. Florence.
\newblock Socratic models: Composing zero-shot multimodal reasoning with language.
\newblock \emph{ArXiv}, abs/2204.00598, 2022.

\bibitem[Zeng et~al.(2021)Zeng, Zhang, and Li]{Zeng2021MultiGrainedVL}
Yan Zeng, Xinsong Zhang, and Hang Li.
\newblock Multi-grained vision language pre-training: Aligning texts with visual concepts.
\newblock \emph{ArXiv}, abs/2111.08276, 2021.

\bibitem[Zhang et~al.(2022{\natexlab{a}})Zhang, Li, Liu, Zhang, Su, Zhu, shuan Ni, and yeung Shum]{Zhang2022DINODW}
Hao Zhang, Feng Li, Shilong Liu, Lei Zhang, Hang Su, Jun-Juan Zhu, Lionel~Ming shuan Ni, and Heung yeung Shum.
\newblock Dino: Detr with improved denoising anchor boxes for end-to-end object detection.
\newblock \emph{ArXiv}, abs/2203.03605, 2022{\natexlab{a}}.

\bibitem[Zhang et~al.(2022{\natexlab{b}})Zhang, Zhang, Hu, Chen, Li, Dai, Wang, Yuan, Hwang, and Gao]{Zhang2022GLIPv2UL}
Haotian Zhang, Pengchuan Zhang, Xiaowei Hu, Yen-Chun Chen, Liunian~Harold Li, Xiyang Dai, Lijuan Wang, Lu Yuan, Jenq-Neng Hwang, and Jianfeng Gao.
\newblock Glipv2: Unifying localization and vision-language understanding.
\newblock \emph{ArXiv}, abs/2206.05836, 2022{\natexlab{b}}.

\bibitem[Zhang et~al.(2021{\natexlab{a}})Zhang, Li, Hu, Yang, Zhang, Wang, Choi, and Gao]{Zhang2021VinVLMV}
Pengchuan Zhang, Xiujun Li, Xiaowei Hu, Jianwei Yang, Lei Zhang, Lijuan Wang, Yejin Choi, and Jianfeng Gao.
\newblock Vinvl: Making visual representations matter in vision-language models.
\newblock \emph{ArXiv}, abs/2101.00529, 2021{\natexlab{a}}.

\bibitem[Zhang et~al.(2021{\natexlab{b}})Zhang, Li, Hu, Yang, Zhang, Wang, Choi, and Gao]{zhang2021vinvl}
Pengchuan Zhang, Xiujun Li, Xiaowei Hu, Jianwei Yang, Lei Zhang, Lijuan Wang, Yejin Choi, and Jianfeng Gao.
\newblock Vinvl: Making visual representations matter in vision-language models.
\newblock \emph{CVPR 2021}, 2021{\natexlab{b}}.

\bibitem[Zhang et~al.(2023)Zhang, Shen, Shi, Cai, Fang, Deng, Yang, Modolo, Tu, and Soatto]{zhang2023musketeer}
Zhaoyang Zhang, Yantao Shen, Kunyu Shi, Zhaowei Cai, Jun Fang, Siqi Deng, Hao Yang, Davide Modolo, Zhuowen Tu, and Stefano Soatto.
\newblock Musketeer: Joint training for multi-task vision language model with task explanation prompts.
\newblock \emph{arXiv preprint arXiv:2305.07019}, 2023.

\bibitem[Zhong et~al.(2019)Zhong, Tang, and Yepes]{zhong2019publaynet}
Xu Zhong, Jianbin Tang, and Antonio~Jimeno Yepes.
\newblock Publaynet: largest dataset ever for document layout analysis.
\newblock In \emph{2019 International Conference on Document Analysis and Recognition (ICDAR)}, pages 1015--1022. IEEE, 2019.

\bibitem[Zhou et~al.(2021)Zhou, Yang, Loy, and Liu]{Zhou2021LearningTP}
Kaiyang Zhou, Jingkang Yang, Chen~Change Loy, and Ziwei Liu.
\newblock Learning to prompt for vision-language models.
\newblock \emph{International Journal of Computer Vision}, 130:\penalty0 2337 -- 2348, 2021.

\bibitem[Zhou et~al.(2022)Zhou, Yang, Loy, and Liu]{Zhou2022ConditionalPL}
Kaiyang Zhou, Jingkang Yang, Chen~Change Loy, and Ziwei Liu.
\newblock Conditional prompt learning for vision-language models.
\newblock \emph{2022 IEEE/CVF Conference on Computer Vision and Pattern Recognition (CVPR)}, pages 16795--16804, 2022.

\bibitem[Zhou et~al.(2019)Zhou, Palangi, Zhang, Hu, Corso, and Gao]{Zhou2019UnifiedVP}
Luowei Zhou, Hamid Palangi, Lei Zhang, Houdong Hu, Jason~J. Corso, and Jianfeng Gao.
\newblock Unified vision-language pre-training for image captioning and vqa.
\newblock \emph{ArXiv}, abs/1909.11059, 2019.

\bibitem[Zhuge et~al.(2021)Zhuge, Gao, Fan, Jin, Chen, Zhou, Qiu, and Shao]{Zhuge2021KaleidoBERTVP}
Mingchen Zhuge, Dehong Gao, Deng-Ping Fan, Linbo Jin, Ben Chen, Hao Zhou, Minghui Qiu, and Ling Shao.
\newblock Kaleido-bert: Vision-language pre-training on fashion domain.
\newblock \emph{2021 IEEE/CVF Conference on Computer Vision and Pattern Recognition (CVPR)}, pages 12642--12652, 2021.

\end{thebibliography}
}

\appendix
\clearpage
\setcounter{page}{1}
\maketitlesupplementary

\section{Quantitative Results on Joint Training}
\label{sec:joint}
We provide some additional discussions for the impact of joint training language and location tasks. Directly combining S4$_{NL}$ objectives and S4$_{Loc}$ objectives, which we denote as S4$_{Joint}$, harms the performance on most downstream tasks. For S4$_{Joint}$, we use the following weighted loss as it gives the best average performance across all tasks:\\
\begin{verbatim}
loss_weights = {
    'screen2html': 1.0,
    'attribute_prediction': 0.5,
    'title_generation': 0.5,
    'node_relation_prediction':0.1,
    'table_parsing':0.1,
    'ocr':0.1,
    'table_detection':0.1,
    'layout_analysis':0.1,
    'image_grounding':0.1,
    'element_grounding':0.1
}
\end{verbatim}

% \begin{table*}
\centering
\begin{tabular}{l|ccc|ccccc}
\toprule
  \multirow{2}{*}{\bf Methods} & \bf Pre-training & \bf Pre-training& \bf Finetune & \multirow{2}{*}{\bf ChartQA $\uparrow$ } & \bf \multirow{2}{*}{\refcls $\uparrow$} & \bf Widget $\uparrow$ & \bf Screen $\uparrow$ &  \multirow{2}{*}{\bf WebSRC $\uparrow$} \\
  &\bf Dataset & \bf Objectives & \bf Batchsize &  &  & \bf Cap. & \bf Sum.\\
% \hline
\midrule

S4\textsuperscript{*}  \textbf{(Ours)}& S4 Data - 15M & S4$_{NL}$ & 8 & \textbf{55.0} &  \textbf{94.9} & \textbf{130.6} & \textbf{105.7} & \textbf{61.1} \\

S4\textsuperscript{*}  \textbf{(Ours)}& S4 Data - 15M & S4$_{Joint}$ & 8 & 52.1 &  \textbf{94.9}  & 128.4 & 101.5 & 60.3 \\

\bottomrule

\end{tabular}

% \resizebox{0.9\linewidth}{!}{

\begin{tabular}{l|cc|cccc}
\toprule
  \multirow{3}{*}{\bf Methods} &  \multirow{3}{*}{\begin{tabular}{@{}c@{}} \bf Pre-training \\ \bf Dataset\end{tabular}}   & \multirow{3}{*}{\begin{tabular}{@{}c@{}} \bf Pre-training \\ \bf Objectives\end{tabular}} & \bf RefExp $\uparrow$ & \bf PublayNet $\uparrow$ & \bf PubTables1M $\uparrow$ & \bf ICDAR 2019 $\uparrow$ \\
  % \hline
  & & &$_{\mathrm{cand\_free}}$ & & \bf (Table Det.) & \bf (Modern Subset)\\
 &&& \bf 30k samples & \bf 1M samples & \bf 400k samples & 
 \bf 600 samples \\

\midrule

S4\textsuperscript{*} (\textbf{Ours}) & S4 Data - 15M& S4$_{Loc}$ & \textbf{84.3}  & \textbf{93.1 }& \textbf{99.0} &  \textbf{79.4} \\

S4\textsuperscript{*} (\textbf{Ours}) & S4 Data - 15M& S4$_{Joint}$ & 79.2  & 91.6 & 97.7 & 76.7 \\
\bottomrule
\end{tabular}
% }

% \end{table*}

\section{Qualitative Results during Pre-training}
We visualize qualitative pre-training results in below figures. For table detection, image grounding, and element grounding, red boxes denotes ground truth boxes and blue boxes denotes predicted boxes. We present four images for each of these three tasks, where the bottom right is a failure case and the others are good cases. For layout analysis, we present a pair of prediction (blue) and ground truth layout (red).

\begin{figure*}
\centering
 \includegraphics[width=\textwidth]{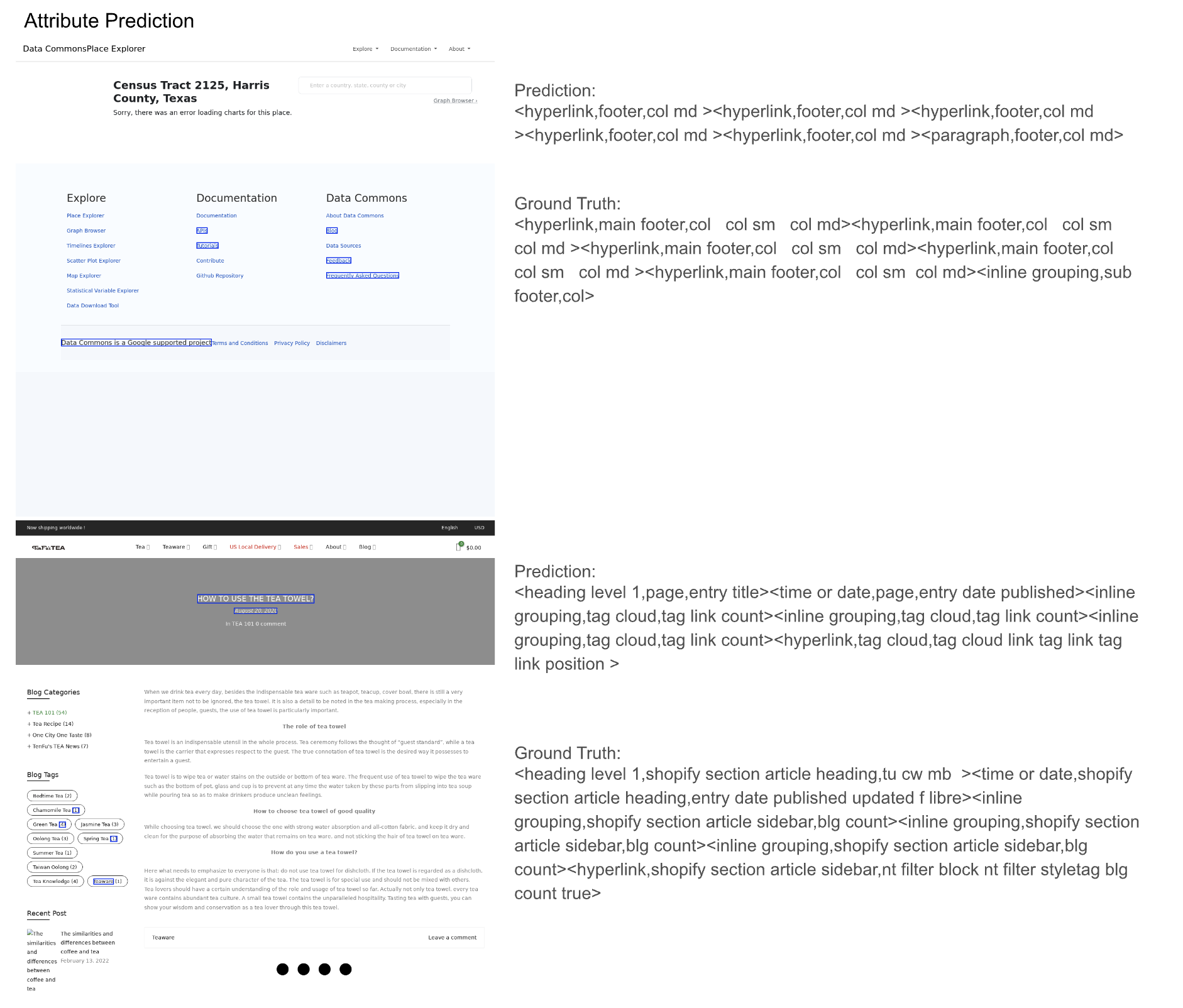}
\caption{Attribute Prediction} 
\end{figure*}

\begin{figure*}
\centering
 \includegraphics[width=\textwidth]{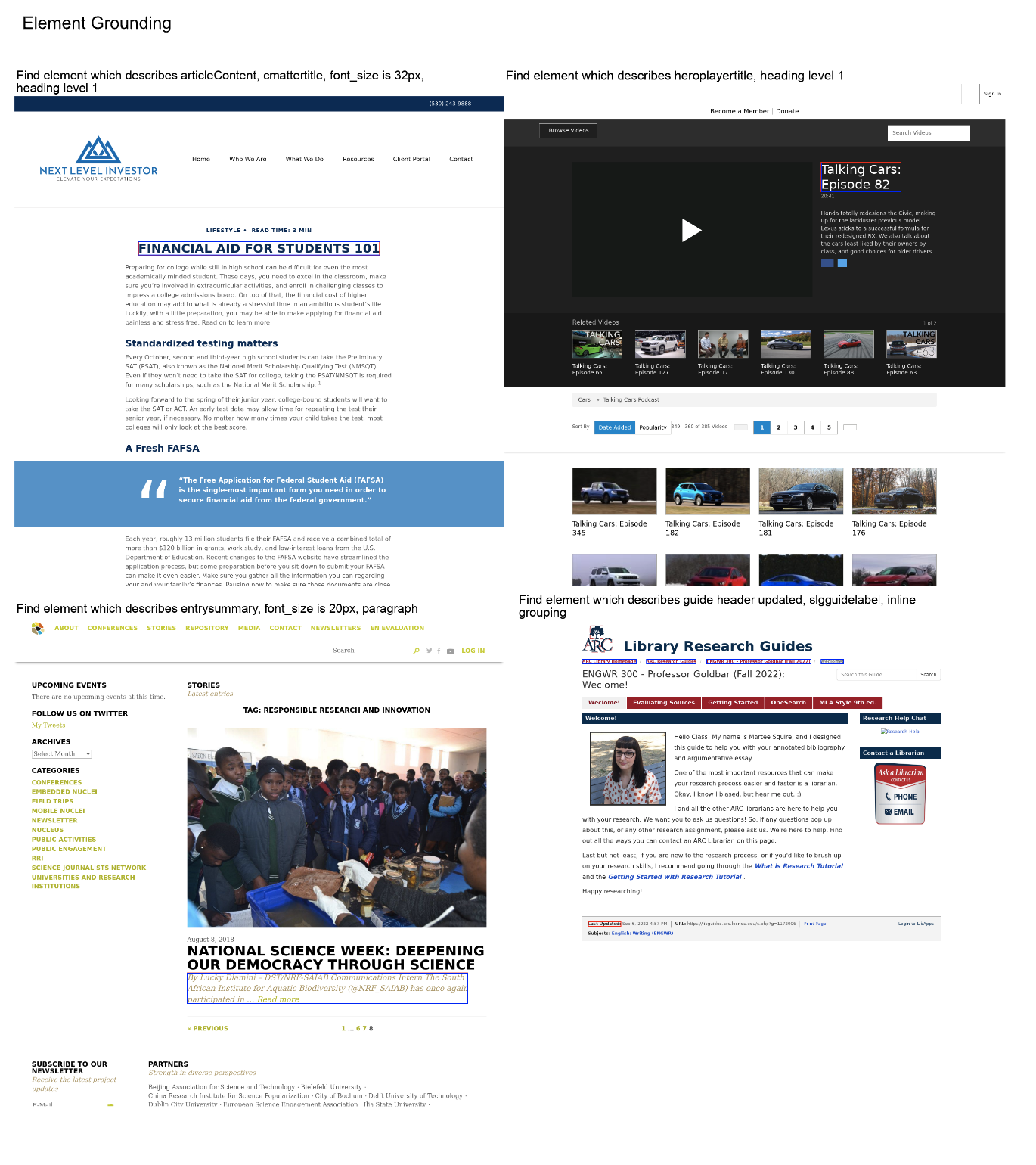}
\caption{Element Grounding} 
\end{figure*}

\begin{figure*}
\centering
 \includegraphics[width=\textwidth]{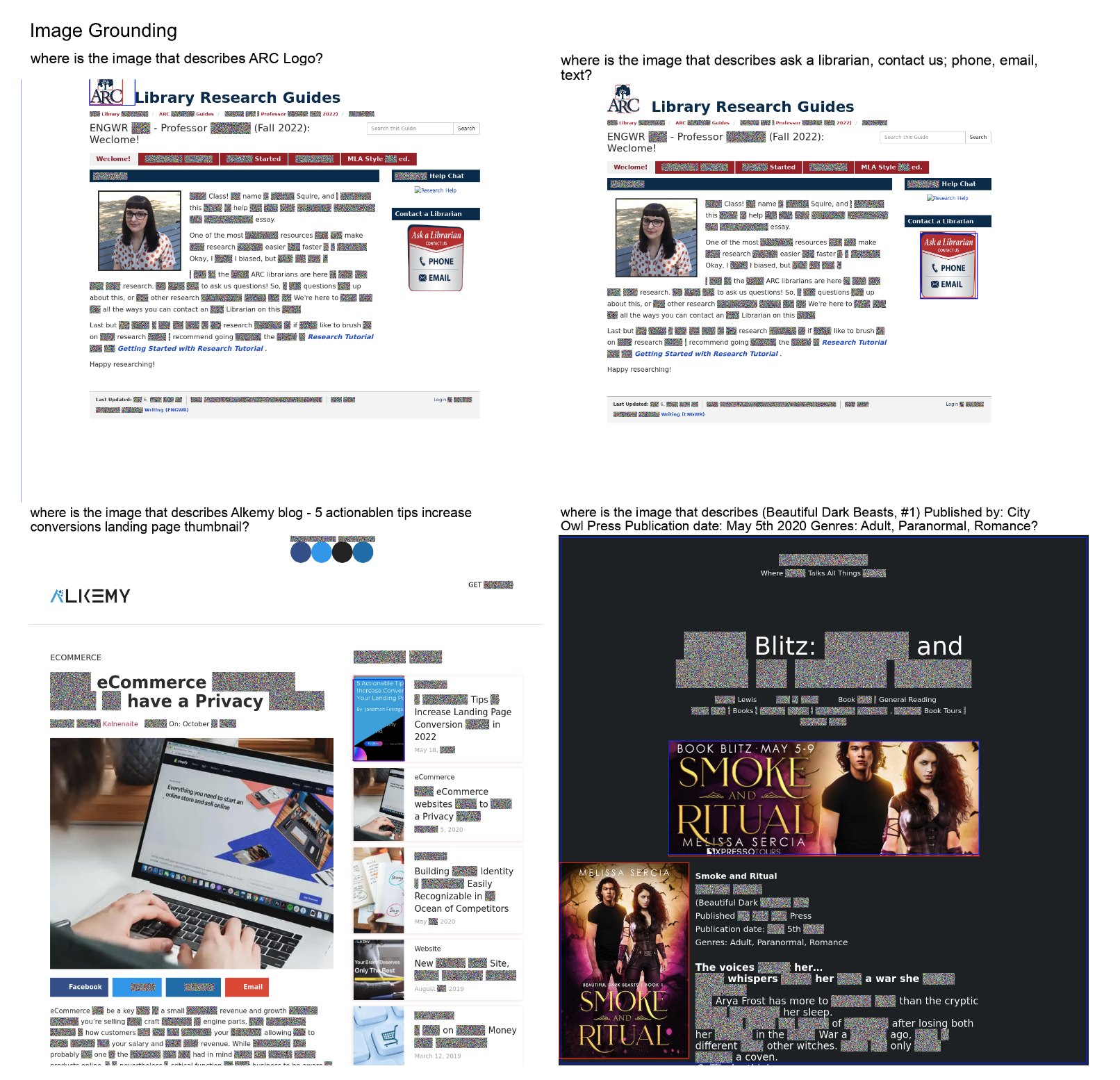}
\caption{Image Grounding} 
\end{figure*}

\begin{figure*}
\centering
 \includegraphics[width=\textwidth]{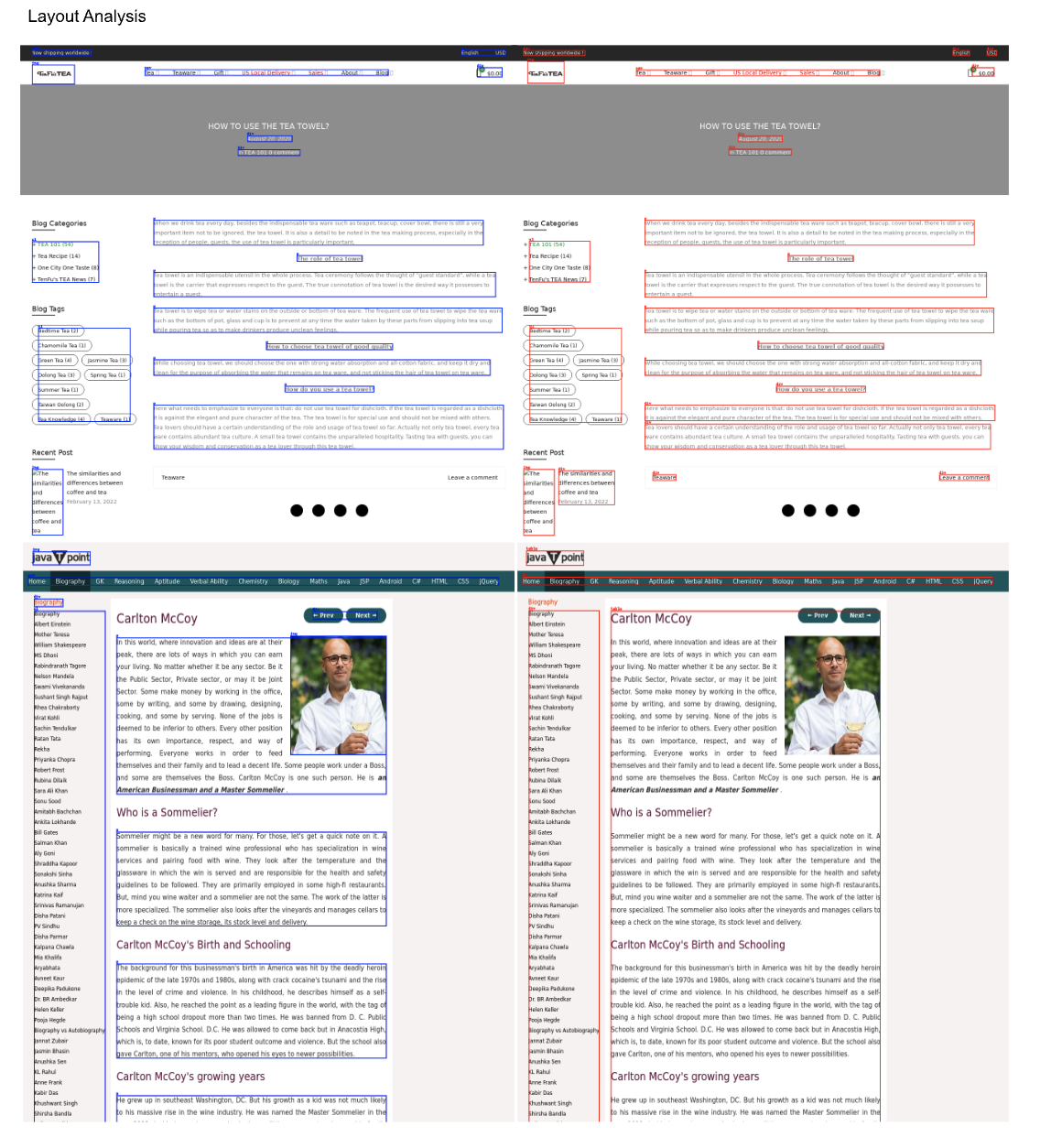}
\caption{Layout Analysis} 
\end{figure*}

\begin{figure*}
\centering
 \includegraphics[width=0.7\textwidth]{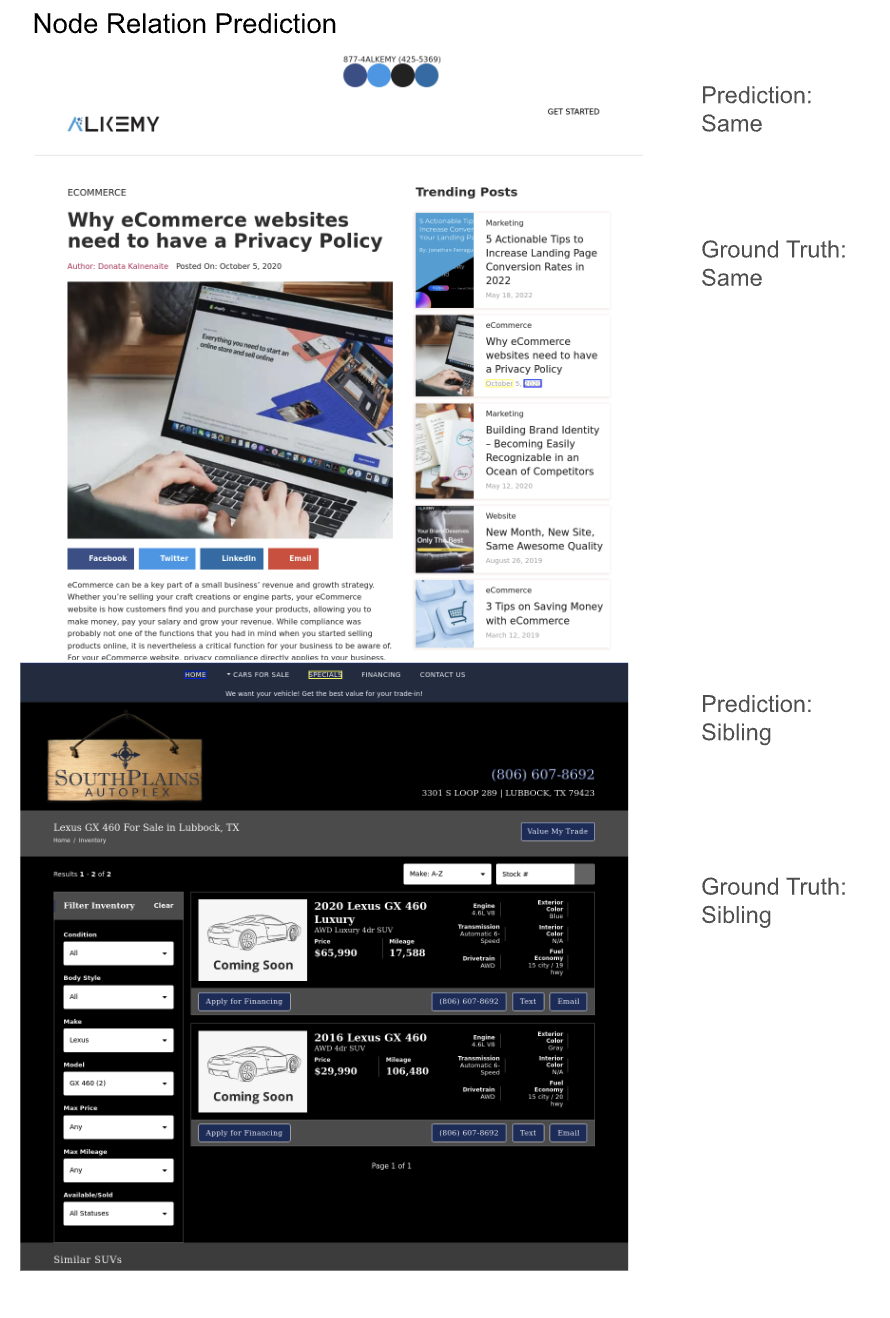}
\caption{Node Relation} 
\end{figure*}

\begin{figure*}
\centering
 \includegraphics[width=\textwidth]{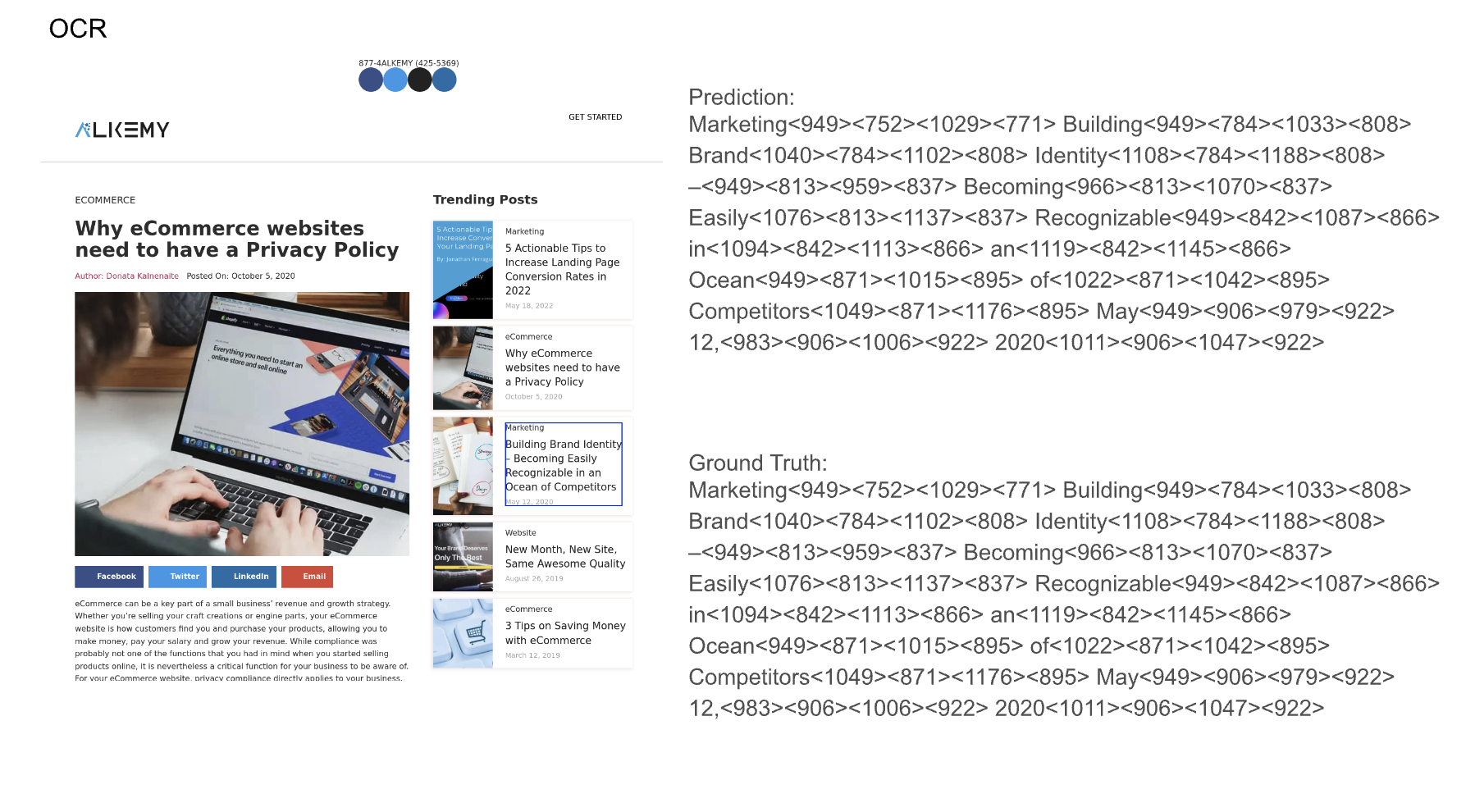}
\caption{OCR} 
\end{figure*}

\begin{figure*}
\centering
 \includegraphics[width=\textwidth]{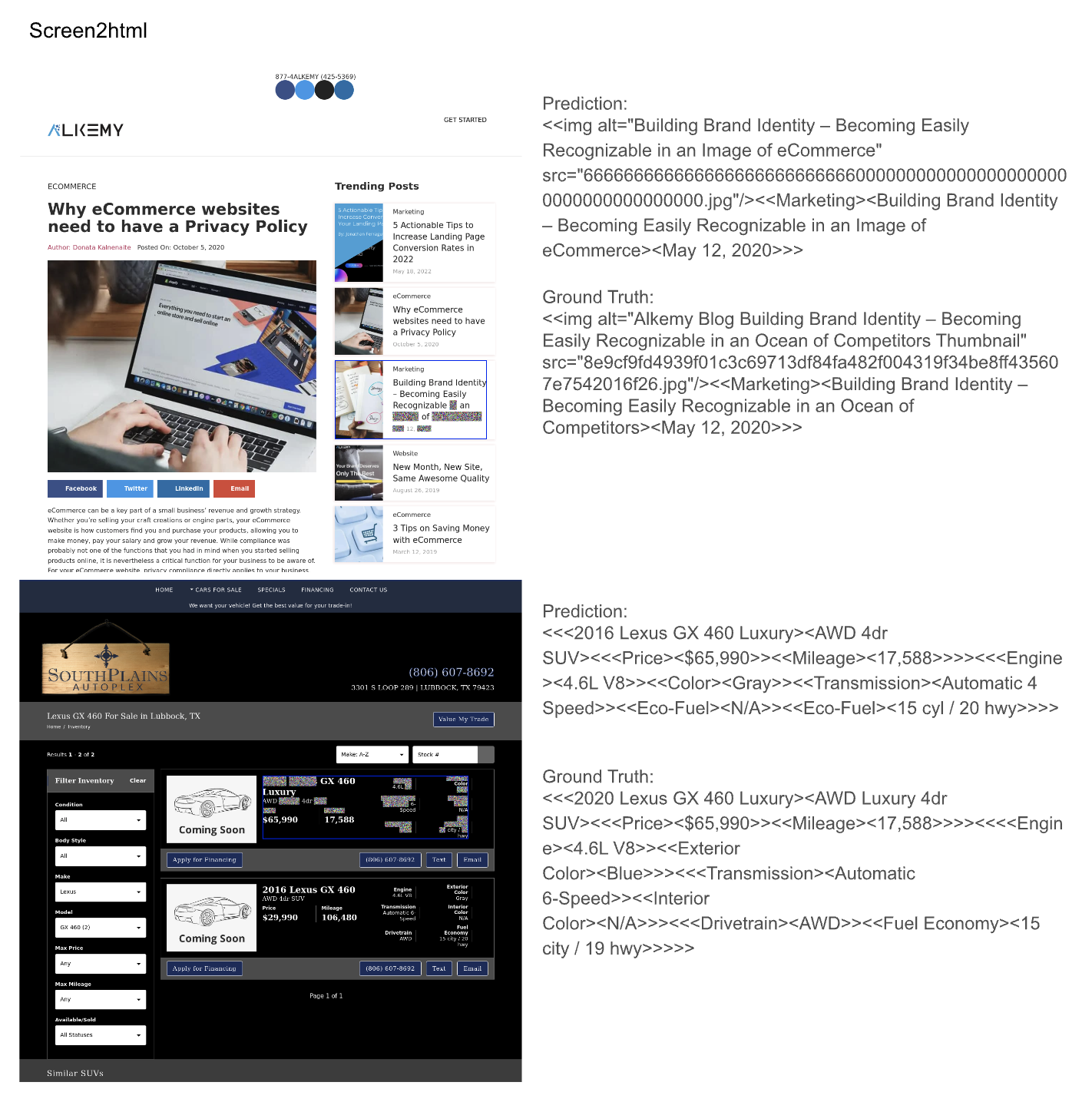}
\caption{Screen2html} 
\end{figure*}

\begin{figure*}
\centering
  \includegraphics[width=\textwidth]{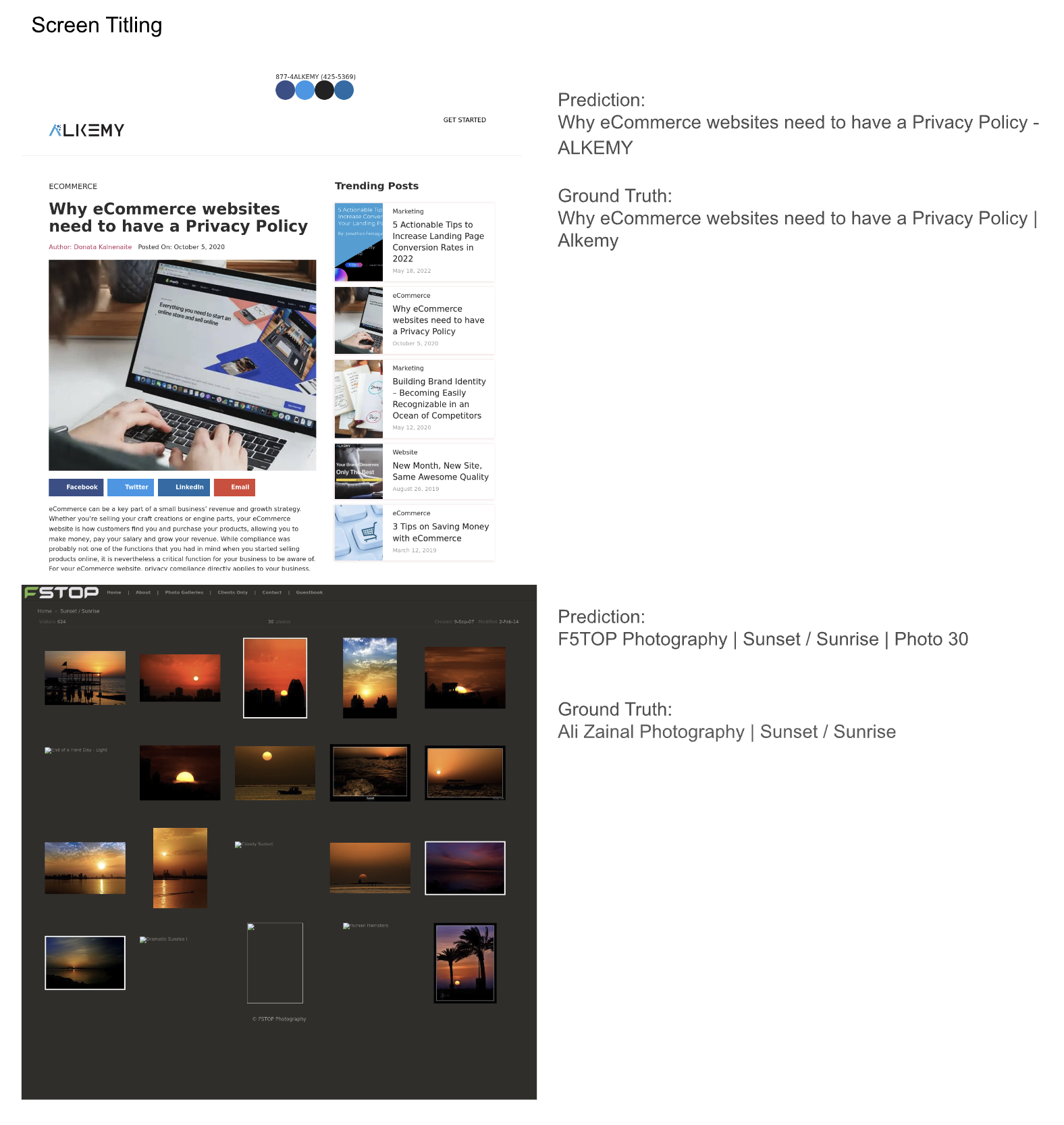}
\caption{Screen Tiltling} 
\end{figure*}

\begin{figure*}
\centering
 \includegraphics[width=\textwidth]{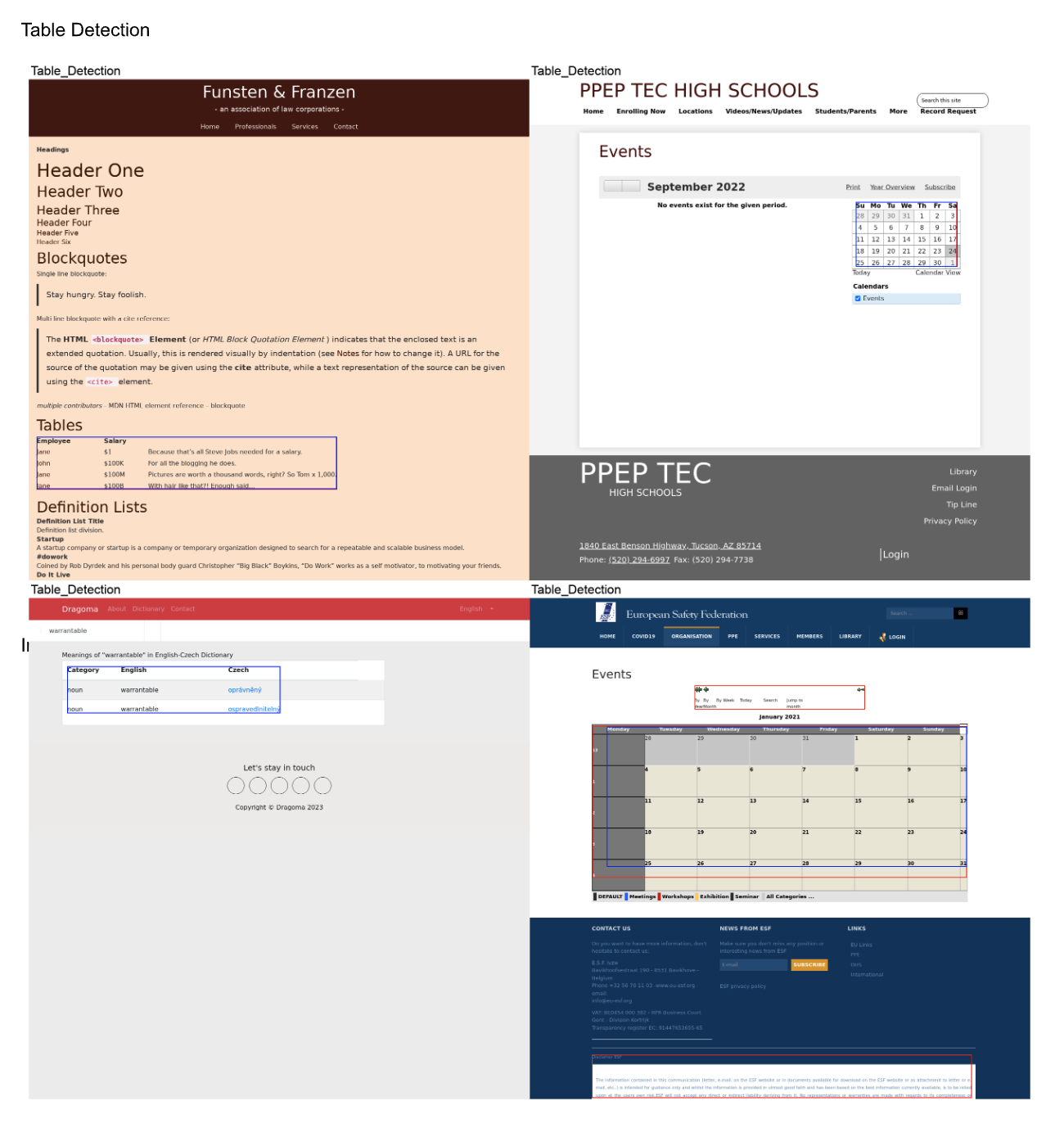}
\caption{Table Detection} 
\end{figure*}

\begin{figure*}
\centering
 \includegraphics[width=\textwidth]{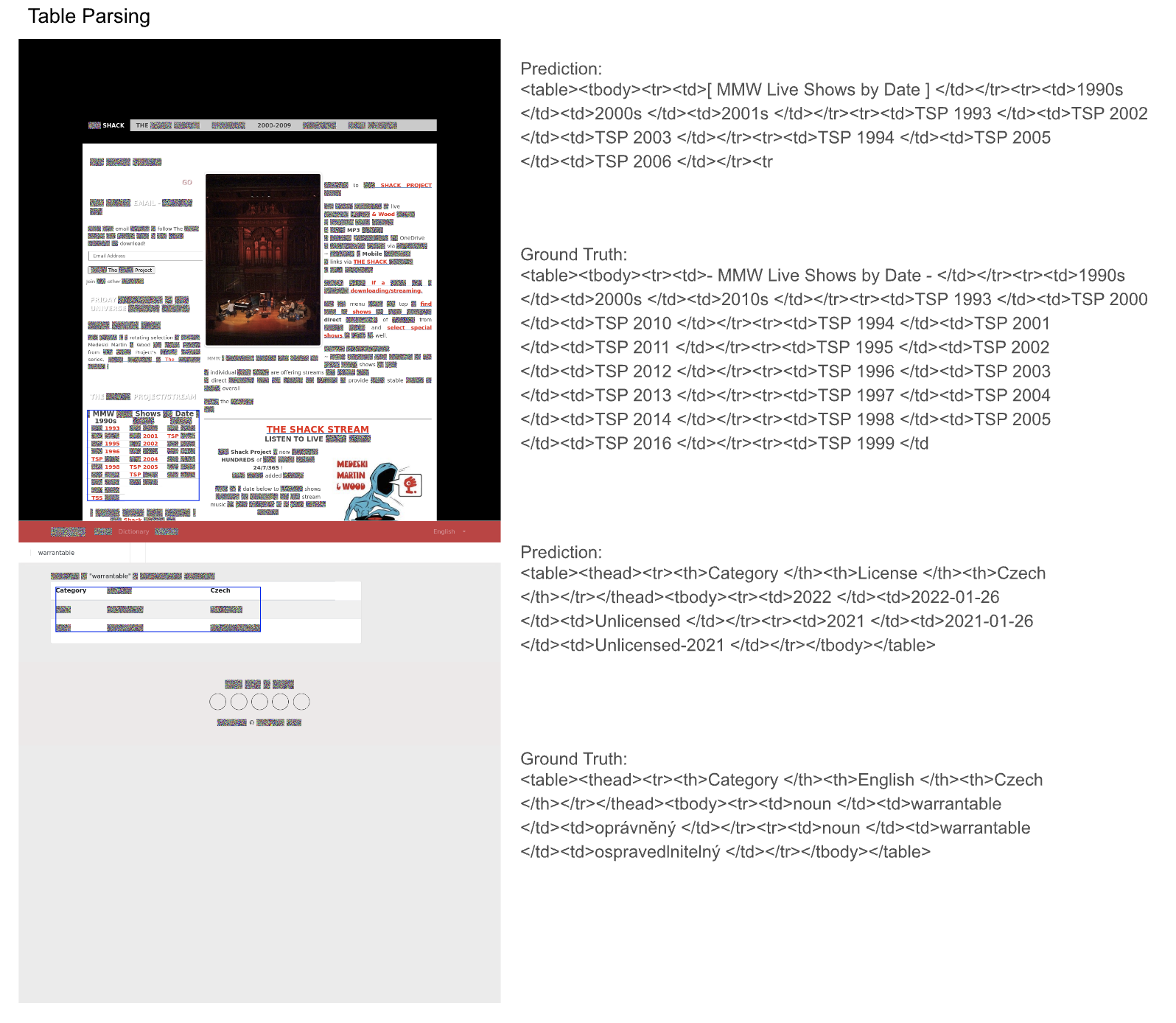}
\caption{Table Parsing} 
\end{figure*}

\end{document}